\documentclass{article}

\usepackage[final]{neurips_2025}

\usepackage[utf8]{inputenc} 
\usepackage[T1]{fontenc}    
\usepackage{hyperref}       
\usepackage{url}            
\usepackage{booktabs}       
\usepackage{amsfonts}       
\usepackage{nicefrac}       
\usepackage{microtype}      
\usepackage{xcolor}         
\usepackage{graphicx}  
\usepackage{subcaption} 

\usepackage{listings}
\usepackage[scaled=0.9]{inconsolata} 
\usepackage{capt-of}
\usepackage{float}

\usepackage{xcolor}

\title{GDS Agent for Advanced Graph Algorithmic Reasoning}

\author{%
  Borun Shi\\
  \texttt{brian.shi@neo4j.com} \\
  \And
  Lusine Adunts \\
  \texttt{lusine.adunts@neo4j.com} \\
  \And
  Ioannis Panagiotas\\
  \texttt{ioannis.panagiotas@neo4j.com} \\
}

\begin{document}

\maketitle

\begin{center}
    \vspace{-2em}
  \includegraphics[width=0.25\textwidth]{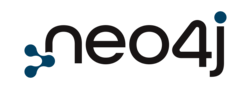}
\end{center}

\begin{abstract}
Large language models (LLMs) have shown remarkable multimodal information processing and reasoning ability. When equipped with tools through function calling and enhanced with retrieval-augmented techniques, compound LLM-based systems can access closed data sources and answer questions about them. However, they still struggle to process and reason over large-scale graph-structure data.
In this technical report we introduce the GDS (Graph Data Science) agent.
The GDS agent introduces a comprehensive set of graph algorithms as tools, together with preprocessing (retrieval) and postprocessing of algorithm results, in a model context protocol (MCP) server. It is accompanied by a skill that guides the LLM through graph analysis.
The agent can be used with any modern LLM out-of-the-box. 
GDS agent allows users to ask any question that implicitly and intrinsically requires graph algorithmic reasoning about their data, and quickly obtain accurate and grounded answers. 
We introduce new benchmarks that evaluate intermediate tool calls as well as final responses, and the results indicate that the GDS agent solves a wide spectrum of graph tasks. 
We also demonstrate how the GDS agent can perform open-ended analyses at the level of domain experts, on tasks such as money laundering tracing and pattern mining in biology. Finally, we discuss the remaining challenges and the future roadmap.

\center \url{https://github.com/neo4j-contrib/gds-agent}

\end{abstract}

\section{Introduction}
Recent progress in transformer-based large language models (LLMs) has shown remarkable reasoning and tool-use ability~\citep{openai2026gpt55, openai2026gpt56sol, anthropic2026fable51, anthropic2026opus55, anthropic2026sonnet55, openai2025gpt5, 2025gemini, guo2025deepseek, yang2025qwen3, openai2024gpt4o, grattafiori2024llama}. However, a fundamental limitation of these models is that they cannot directly access and interact with private data. Retrieval-augmented methods~\citep{lewis2020retrieval} address the problem by retrieving relevant content from a private database for a user question and then feeding it into an LLM for reasoning. Function calling~\citep{functiona-calling, structured-output} capabilities allow an LLM to generate structured outputs which are used to invoke external tool calling. Such a compound system~\citep{compound-ai-blog} of LLMs and tools (other software libraries and APIs) becomes an \textit{agent}. An agent~\citep{yao2023react} is triggered by a user question (prompt), driven by the LLMs which call tools to interact with external systems and databases. There are a growing number of agents for a variety of use cases, including being a coding assistant, domain-expert researcher, and medical assistant~\citep{plaat2025agentic}.

Among many modalities of data, graphs naturally represent a wide range of real-world data. A social network can be modeled as a graph with users being the nodes and interactions between the users forming the edge set. 
An underground transport network has stations as nodes and connections as edges. Records of financial transactions can be modeled as a graph with accounts as the nodes and transactions as the edges.
Many real-world graphs are heterogeneous, containing different node and edge types, and attributed, having diverse node and edge properties of different types. Such graphs are commonly referred to as Knowledge Graphs (KGs).

Current LLMs, including reasoning models, have limited graph reasoning ability, especially on large graphs and when the answer depends on how the graph is presented~\citep{taylor2024large, zhang2026exposing, xu2026gta}. Existing theoretical and empirical studies either tokenize individual nodes and edges for transformers~\citep{sanford2024understanding, kim2022pure} or textualize the graph and prompt the LLM~\citep{fatemi2023talk}, sometimes after retrieving a subgraph~\citep{he2024g}. The first setup limits out-of-the-box adoption with existing LLMs. The second depends on an ad hoc textualization scheme, is bounded by the context window and inherits issues that arise from long context~\citep{liu2023lost}.

Existing benchmarks fall into three categories. The first is knowledge-base question answering~\citep{yang2018hotpotqa, wu24stark}, which represents few-hop retrieval on KGs. For example, the question ``Who is a friend of Alice that knows Bob?'' might have answer ``Charlie'' in a social network database and can be answered by matching a short pattern, without running any graph algorithm. The second category measures how well LLMs are able to simulate various graph algorithms~\citep{wang2023nlgraph, tang2024grapharena, xu2026graphomni}. The questions are commonly given by textualized graphs, such as ``My graph is represented by an edge list: [(0,1),(1,2),(2,3)]. How many triangles are there?''. These benchmarks mostly use basic algorithms, such as breadth-first search and shortest paths, on small graphs~\citep{markeeva2024clrs, taylor2024large} or on long edge lists that fill the context window~\citep{graphwalks}. More recent ones use harder problems and larger graphs. In this case, the reasoning models lose accuracy as the graphs grow~\citep{zhang2026exposing}. One recent benchmark gives the model summary statistics computed from random-walk samples instead of the graph itself~\citep{maurya2026evaluating}. 

The third category gives LLM agents executable graph tools or code. ProGraph asks the model to write code against graph libraries such as NetworkX~\citep{li2024professionals}. GraphCogent uses a multi-agent framework with tools on real-world graphs~\citep{wang2025graphcogent}, and GABench provides tools on real-world graphs and scores both the tool selection and the final answer; it reports that agents still struggle on complex tasks and that the quality of tool calls matters more than their number~\citep{tan2026gabench}. GraphWalk gives the agent basic traversal operations on synthetic graphs and the agent with these tools stays accurate as the graphs grow while the one without them does not~\citep{ghandi2026graphwalk}.

For graph databases, existing work focuses on retrieval and querying. NGDBench evaluates Text-to-Cypher and GraphRAG on noisy graphs with pattern-matching and aggregation queries and finds sensitivity to noise and weak analytical reasoning~\citep{yufei2026ngdbench}. GraphRAFT generates Cypher queries to retrieve subgraphs on STaRK~\citep{clemedtson2025graphraft}. GraphSeek lets the LLM plan predefined graph operations instead of writing queries and reports higher success than direct query generation~\citep{besta2026graphseek}. None of these tasks requires graph algorithms, which query languages alone cannot express in general.

In this technical report, we introduce the GDS~(Graph Data Science) agent, an LLM-based agent equipped with a comprehensive set of graph algorithms as tools and a skill that guides it through graph analysis. \textbf{The GDS agent is able to answer a wide range of questions that implicitly and intrinsically require graph algorithmic reasoning on real-world knowledge graphs}.

We evaluate the agent with new benchmarks of the third category (Section~\ref{sec:benchmarks}). The tools include graph algorithms, retrieval of subgraphs from the database, and obtaining graph schema from the DB. We check both the expected tool calls with their parameters and the final answers.

We evaluate the agent on open-ended tasks with real data. For the 2025 Bybit theft database~\citep{fbi2025bybit}, we ask the agent to find laundering addresses. GDS agent is able to find the majority of those published by the FBI. On the FlyWire fly-brain connectome database, one general request for a structural analysis recovers statistics of a published study~\citep{lin2024network}.

GDS agent can act as a helpful assistant for a wide range of complex analysis tasks on large-scale KGs, from answering a specific question to carrying out an open-ended analysis. Users are able to collaborate with the agent to
analyze their graphs. This amplifies the usefulness of LLMs for large-scale KGs and removes the barrier of leveraging graph analytics libraries.

\section{Preliminaries}
\begin{figure}
  \centering
  \includegraphics[width=0.98\textwidth]{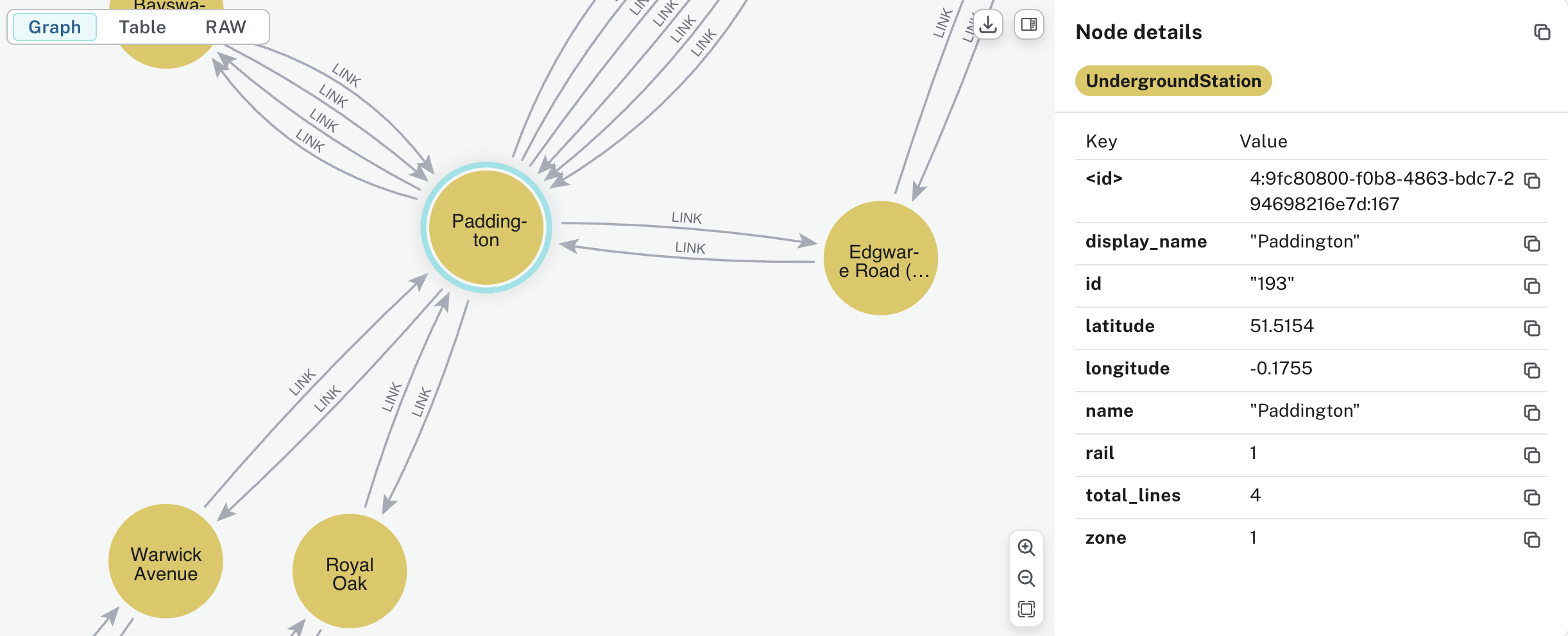}
  \caption{A snapshot of an example KG stored in Neo4j, representing a version of the London underground map. The nodes represent stations, each having textual and numerical properties such as name and total\_lines. The edges represent station links, each having properties such as line, distance and time.}
  \label{fig:ln-map}
\end{figure}

Many real-world KGs are stored in native graph databases such as Neo4j~\citep{neo4j} which offers efficient disk storage and access. An example of such graph is shown in Figure~\ref{fig:ln-map}. Users are able to query their data with Cypher~\citep{cypher}. Other similar graph query languages include openCypher~\citep{opencypher}. They evolved into the recently formed GQL~\citep{gql} standard. The key feature of Cypher, and all graph query languages, is graph pattern matching. As a basic example, \texttt{MATCH (n:A)-[:R1]->(m:B) RETURN DISTINCT m} finds all distinct nodes of type \texttt{B} that are some target nodes of relationships with type \texttt{R1}, from any source node of type \texttt{A}. Query engines built on top of graph databases are able to optimize query plans, ensuring efficient execution. The expressiveness of the query language~\citep{gheerbrant2024gql} ensures a wide range of end-user questions can be formulated as Cypher queries.

However, there still exist a wide range of real-world questions that cannot be answered with Cypher alone. For example, finding various shortest paths, computing centrality scores and detecting communities of nodes under some optimization criteria. Such tasks correspond to running graph algorithms on KGs. The GDS~\citep{gds} library offers a comprehensive set of such algorithms. The available algorithms range from centrality, path-finding, community to node similarity algorithms. The implementations are highly parallelizable and memory-efficient.

In order to execute a GDS algorithm, one has to first perform a \textit{Cypher projection} which constructs an in-memory (sub)graph from the Neo4j database. In-memory projection provides subgraph retrieval with property and type filtering, as well as on-the-fly transformations. String properties present in the database are not projected, since none of the classical graph algorithms leverages such property. Multiple algorithms can be executed on a projected graph and the projected graph can also be mutated, for example with additional properties. The actual algorithms can run efficiently on the projected graph, independent of the database. An example Cypher projection is given in Figure~\ref{fig:cypher-projection}.

\begin{figure}
\centering
\begin{minipage}{0.98\linewidth}
\begin{lstlisting}
MATCH (n)-[r]->(m) WITH n, r, m
RETURN gds.graph.project(
graph_name,
n,
m,
{sourceNodeLabels: labels(n), 
targetNodeLabels: labels(m), 
relationshipType: type(r), 
sourceNodeProperties: {zone: toFloat(n.zone), total_lines: n.total_lines},
targetNodeProperties: {zone: toFloat(m.zone), total_lines: m.total_lines}, 
relationshipProperties: {distance: r.distance, time: r.time})
\end{lstlisting}
\end{minipage}
\caption{An example Cypher projection on the database in ~Figure~\ref{fig:ln-map}. The Cypher query first retrieves all nodes and directed edges. \texttt{WITH} the materialised graph from the DB, an in-memory graph of the name graph\_name is created. Labels on source nodes and target nodes are kept. Types of all relationship are kept. All source nodes have two properties: zone and total\_lines, using values from the DB with appropriate types. All target nodes have the same properties. All relationships have two properties: distance and time. Note that more complex Cypher queries with filtering and aggregations can be used as well.}
\label{fig:cypher-projection}
\end{figure}

\section{Architecture}

\begin{figure}[ht]
  \centering
  \includegraphics[width=\textwidth]{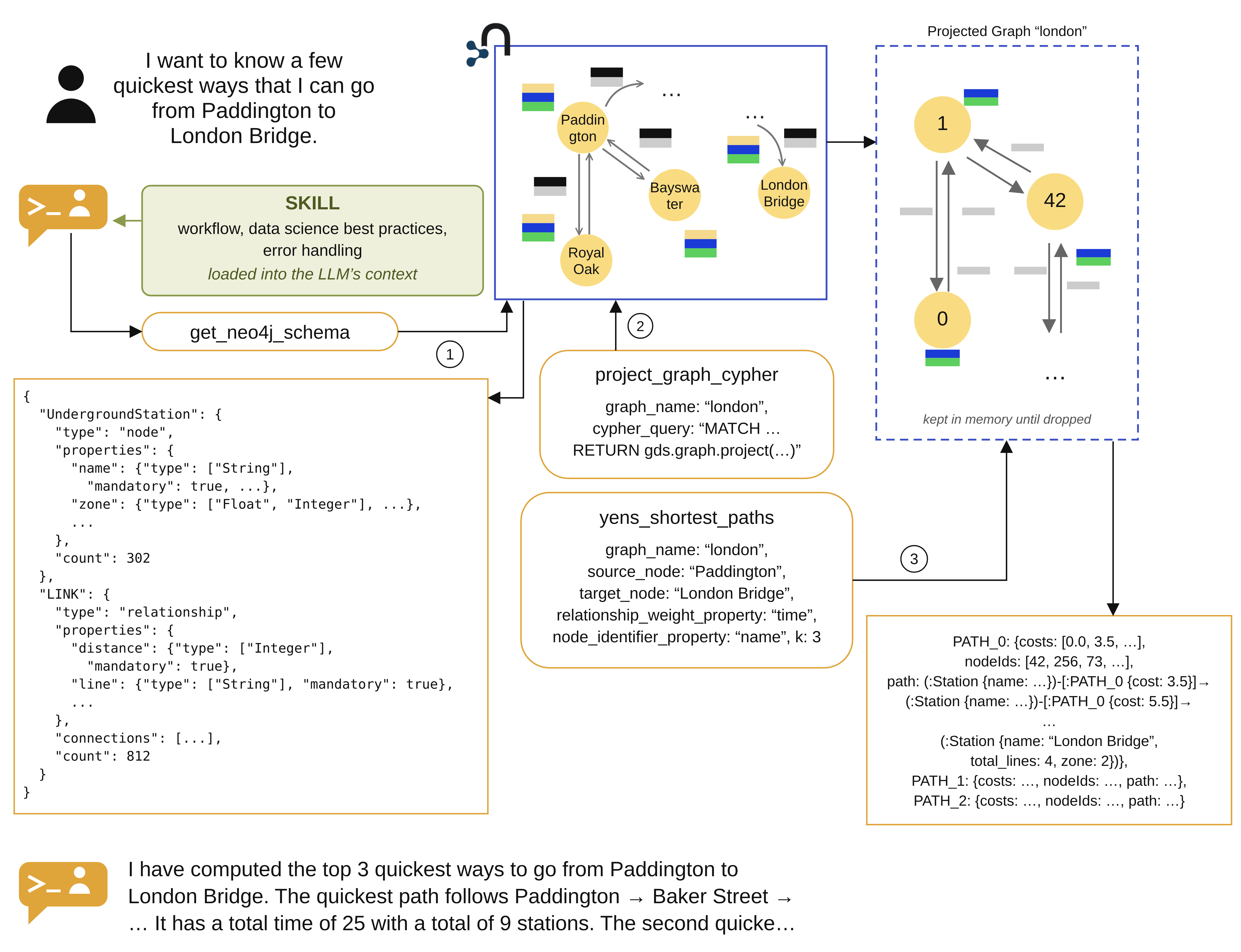}
  \caption{Architecture of GDS Agent by example. A user asks a question about their database shown in Figure~\ref{fig:ln-map}. The GDS agent is equipped with a skill that guides its workflow and invokes several tools to obtain a final answer. Round orange boxes represent tool calls. Square orange boxes represent raw returned data from tool calls. The sequence of actions is given by the numbers. The LLM generates the final answer based on the user question and all tool call results as context.}
  \label{fig:architecture}
\end{figure}

We present the architecture of GDS Agent with a concrete example. The core part of the agent is a Model Context Protocol (MCP) server~\citep{mcp} containing tools including GDS algorithms. Any LLM that supports function calling using structured output may be used as the MCP client within the agent. The server runs the algorithms in one of two modes. In plugin mode, GDS is installed as a plugin in a self-managed Neo4j database. In session mode, the algorithms run in GDS Aura Graph Analytics sessions on AuraDB and the server detects the mode at startup. Session mode adds three tools to create, list and delete sessions. The full list of tools is given in Appendix~\ref{ap:tool-list}. The server is accompanied by a skill, a set of instructions in the LLM's context that describes the typical workflow, data science best practices and how to handle errors. 

For example, a user can ask ``I want to know a few quickest ways that I can go from Paddington to London Bridge'' against a database containing a version of the London underground map from Figure~\ref{fig:ln-map}. Intuitively, the LLM understands that the question requires finding some shortest paths between the source and the target station with the constraints that they need to be ``quickest'' and the requirement of finding ``a few'' of them. The skill is already part of the LLM's context when the question arrives.

\textcircled{\raisebox{-0.9pt}{1}} The agent decides to invoke the \texttt{get\_neo4j\_schema} tool, which returns the node labels, the relationship types and their properties with types. 
The call returned a list of node properties including name (string), id (string), total\_lines (int) and zone (float), and relationship properties line (string), distance (int) and time (int).

\textcircled{\raisebox{-0.9pt}{2}} Each tool in the server has a tool specification using the JSON format in the MCP standard. A tool specification contains a description field, as well as a dictionary of parameters, their types and their descriptions. An example tool specification for Yen's algorithm~\citep{yens} is given in Appendix~\ref{ap:yens}. The LLM inspects the specifications and determines \texttt{yens\_shortest\_paths} as the appropriate tool to call. Algorithms run on an in-memory projected graph, so the LLM first calls the \texttt{project\_graph\_cypher} tool with a name for the graph and a Cypher query that selects the stations and the connections between them, together with the numeric relationship properties \texttt{time} and \texttt{distance}. The projected graph is stored in memory under the given name and persists until it is dropped, so that later calls can reuse it. Note that the projection carries only numeric properties; any textual properties on nodes and edges are not part of it. 

\textcircled{\raisebox{-0.9pt}{3}} The LLM then calls the \texttt{yens\_shortest\_paths} tool with the name of the projected graph, and the tool executes the GDS Yen's algorithm on it. The LLM also creates a dictionary of parameter key-value pairs that are passed to the tool.
These are usually algorithm related, for example the choice of $k$ for Yen's algorithm. As the projected graph does not contain any string properties such as names, the source and target node need to be mapped to the internal ids representing ''Paddington'' and ''London Bridge''. For this, the LLM must first select the appropriate identifying property from those returned by the \texttt{get\_neo4j\_schema} call. The LLM settles on using ''name'' instead of other alternatives like ''rail'' or ''display\_name''. This is passed to the tool with the
\texttt{node\_identifier\_property: "name"} parameter. The tool then internally invokes \texttt{gds.shortestPath.yens} algorithm with the source and target node ids, with parameters \texttt{relationship\_weight\_property: "time"} and \texttt{k: 3}. k specifies the top k shortest paths to be returned. The LLM in this case interprets ``a few'' as finding 3 such paths. The algorithm returns a dataframe of 4 columns: index, costs, nodeIds, path. Each row represents a path, with the corresponding cost of each edge, the list of node ids traversed in the paths, and a full description of the path data, including names. The result dataframe is serialized to a textual description and fed into the LLM.

The LLM generates the final answer based on the user question and all textualised tool responses as context. It provides a summary of the 3 quickest routes, describing the time and detailed path for each of them. A user can obtain this answer with no graph data science expertise.
A user can evaluate and explain the final answer by inspecting their database, as well as examining tool invocations, along with their parameters and the raw returned data.


\section{Benchmarks}
\label{sec:benchmarks}

We introduce two new benchmarks consisting of representative datasets and questions. The \textit{graph-agent-bench-ln-v1} is based on a version of the London underground map \citep{tube-map} shown in Figure~\ref{fig:ln-map}. \textit{graph-agent-bench-got-v1} is based on a version of the Game of Thrones dataset \citep{got-dataset} containing Kings and Knights belonging to different houses and involved in various battles.

For each dataset, we provide a set of questions of varying levels of difficulty that require specific tool calls to answer. For each question, we provide the expected tool calls, the expected tool parameters and expected answers. All questions are written manually by the authors. The algorithm parameters have been specifically selected to minimize the chance of varying results, but they might still happen for a few questions without a unique answer e.g., a node might actually be classified in a different community between runs or multiple correct projections may create differing results.

We also report a benchmark built from a real incident, the 2025 Bybit theft. Its database holds about 17 million raw Ethereum transfers, and the answer can be checked against an externally published list of 51 laundering addresses. Unlike the two benchmarks above, it consists of a single task and is scored on the final list only.

\begin{figure}[h]
\centering
\begin{minipage}{0.98\linewidth}
\begin{lstlisting}
Execute personalised article rank with damping factor 0.8 on Paddington and I want to know the scores for Paddington & Bayswater. Give answer in the format station:answer. Separate the stations' result with a comma.
Expected Tool Calls: ["mcp__gds-agent__get_neo4j_schema", "mcp__gds-agent__project_graph_cypher", "mcp__gds-agent__article_rank"]
Expected Tool Parameters: {"article_rank": {"damping_factor": 0.8, "source_nodes": ["Paddington"], "node_identifier_property": "name", "nodes": ["Paddington", "Bayswater"]}}
Expected Answers: Paddington: 0.235156, Bayswater: 0.036966

Question: Could you help me find out what are the few quickest ways to go from Paddington to London Bridge? Format the results as path: costs, for example, ["A", "B", "C"]: [0.0, 98.7, 100.1], ["A", "E", "C"]: [0.0, 32.5, 150.0].
Expected Tool Calls: ["mcp__gds-agent__get_neo4j_schema", "mcp__gds-agent__project_graph_cypher", "mcp__gds-agent__yens_shortest_paths"]
Expected Tool Parameters: {"yens_shortest_paths": {"source_node": "Paddington", "target_node": "London Bridge", "k": "<=5", "relationship_weight_property": "time", "node_identifier_property": "name"}}
Expected Answers: ["Paddington"", ""Edgware Road (C)"", ""Baker Street"", ""Bond Street"", ""Green Park"", ""Westminster"", ""Waterloo"", ""Southwark"", ""London Bridge""]: [0.0, 3.0, 6.0, 8.0, 10.0, 13.0, 15.0, 16.0, 18.0],["Paddington"", ""Edgware Road (C)"", ""Baker Street"", ""Bond Street"", ""Green Park"", ""Westminster"", ""Waterloo"", ""Southwark"", ""London Bridge""]: [0.0, 3.0, 6.0, 8.0, 10.0, 13.0, 15.0, 16.0, 18.0],["Paddington"", ""Edgware Road (C)"", ""Baker Street"", ""Bond Street"", ""Green Park"", ""Westminster"", ""Waterloo"", ""Southwark"", ""London Bridge""]: [0.0, 3.0, 6.0, 8.0, 10.0, 13.0, 15.0, 16.0, 18.0],["Paddington"", ""Edgware Road (C)"", ""Baker Street"", ""Bond Street"", ""Green Park"", ""Westminster"", ""Waterloo"", ""Southwark"", ""London Bridge""]: [0.0, 3.0, 6.0, 8.0, 10.0, 13.0, 15.0, 16.0, 18.0],["Paddington"", ""Edgware Road (B)"", ""Marylebone"", ""Baker Street"", ""Bond Street"", ""Green Park"", ""Westminster"", ""Waterloo"", ""Southwark"", ""London Bridge""]: [0.0, 3.0, 5.0, 6.0, 8.0, 10.0, 13.0, 15.0, 16.0, 18.0]

Question: Is it the case that I can go from any station to any other station in this transport network? Answer Yes or No and nothing else.
Expected Tool Calls: ["mcp__gds-agent__get_neo4j_schema", "mcp__gds-agent__project_graph_cypher", "mcp__gds-agent__strongly_connected_components"]
Expected Tool Parameters: {"strongly_connected_components": {}}
Expected Answers: Yes

\end{lstlisting}

\end{minipage}
\caption{Some examples from the \texttt{graph-agent-bench-ln-v1} benchmark. The first example instructs the agent to execute a specific parameterised algorithm. The latter two examples are questions that require the agent to understand the intent and pick appropriate tools and parameters.}
\label{fig:benchmark-basic-example}
\end{figure}

Our full benchmark is at \url{https://github.com/brs96/gds-agent-benchmarks}. 
Some examples from \textit{graph-agent-bench-ln-v1} are shown in Figure~\ref{fig:benchmark-basic-example}.

\subsection{Experimental setup}
All experiments use version 1.2.0 of the GDS agent MCP server with its skill. The skill is a set of instructions loaded into the agent's context. It describes the typical workflow, data science best practices and how to handle typical errors. Unless stated otherwise, the LLM used is Claude Sonnet 5.5. For all models that support it we use reasoning effort \textit{medium} and verbosity \textit{low}; GPT-4o has no reasoning effort setting. All other system and model settings, such as temperature and max token limits, are default values. The Claude models are run through the Claude Code command-line client, and the OpenAI models through the OpenAI Agents SDK. Each question has a timeout of 25 minutes.
More results are included in Appendix~\ref{ap:got}.

\subsection{Results on the tool-call benchmarks}
\begin{figure}[h]
\centering
\begin{subfigure}{0.49\textwidth}
    \includegraphics[width=1\linewidth]{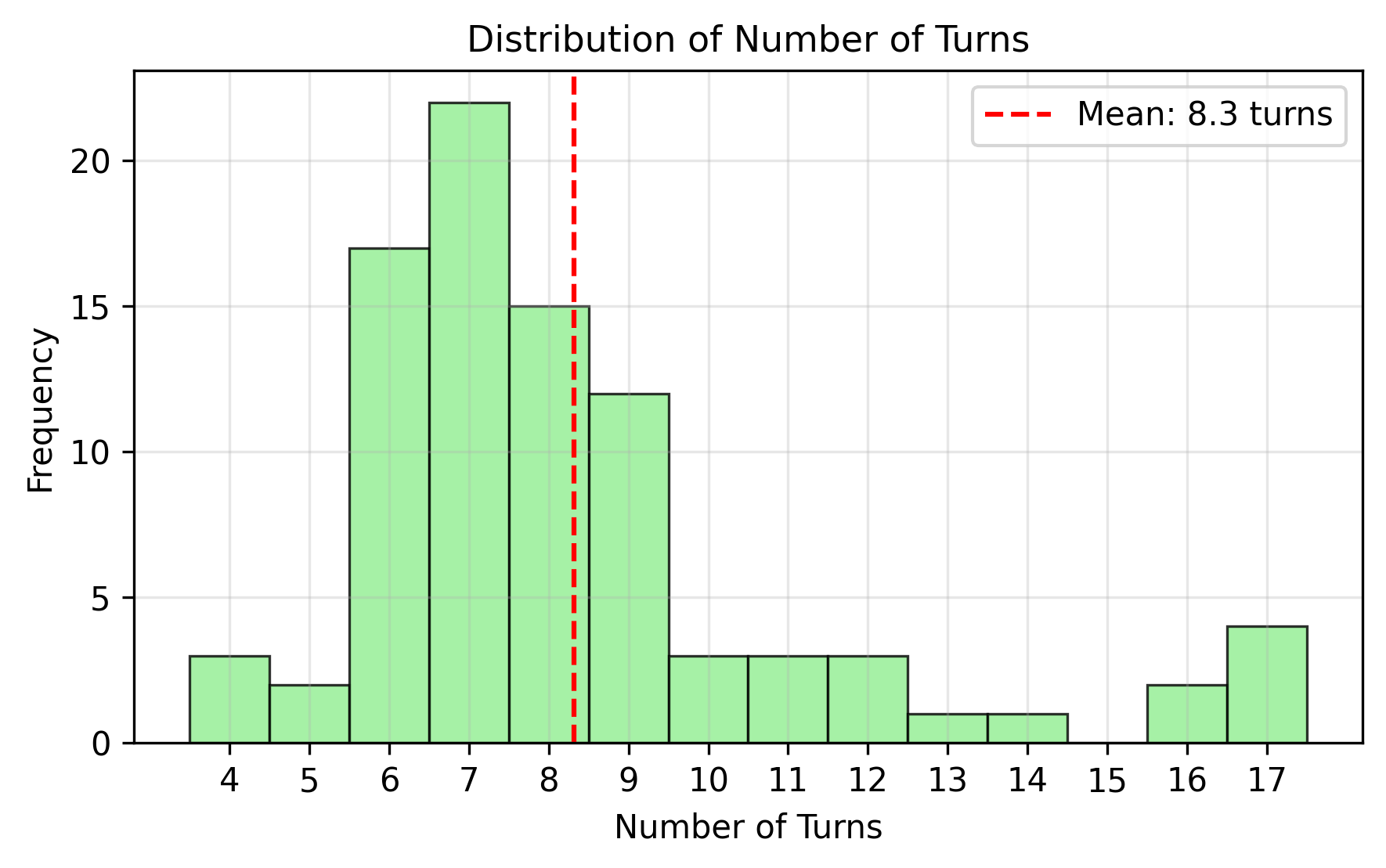}
    \caption{Distribution of number of turns for each question. A tool call, tool response and interaction with the user count towards a turn.}
    \label{fig:turns-distribution}
\end{subfigure}
\hfill
\begin{subfigure}{0.49\textwidth}
    \includegraphics[width=1\linewidth]{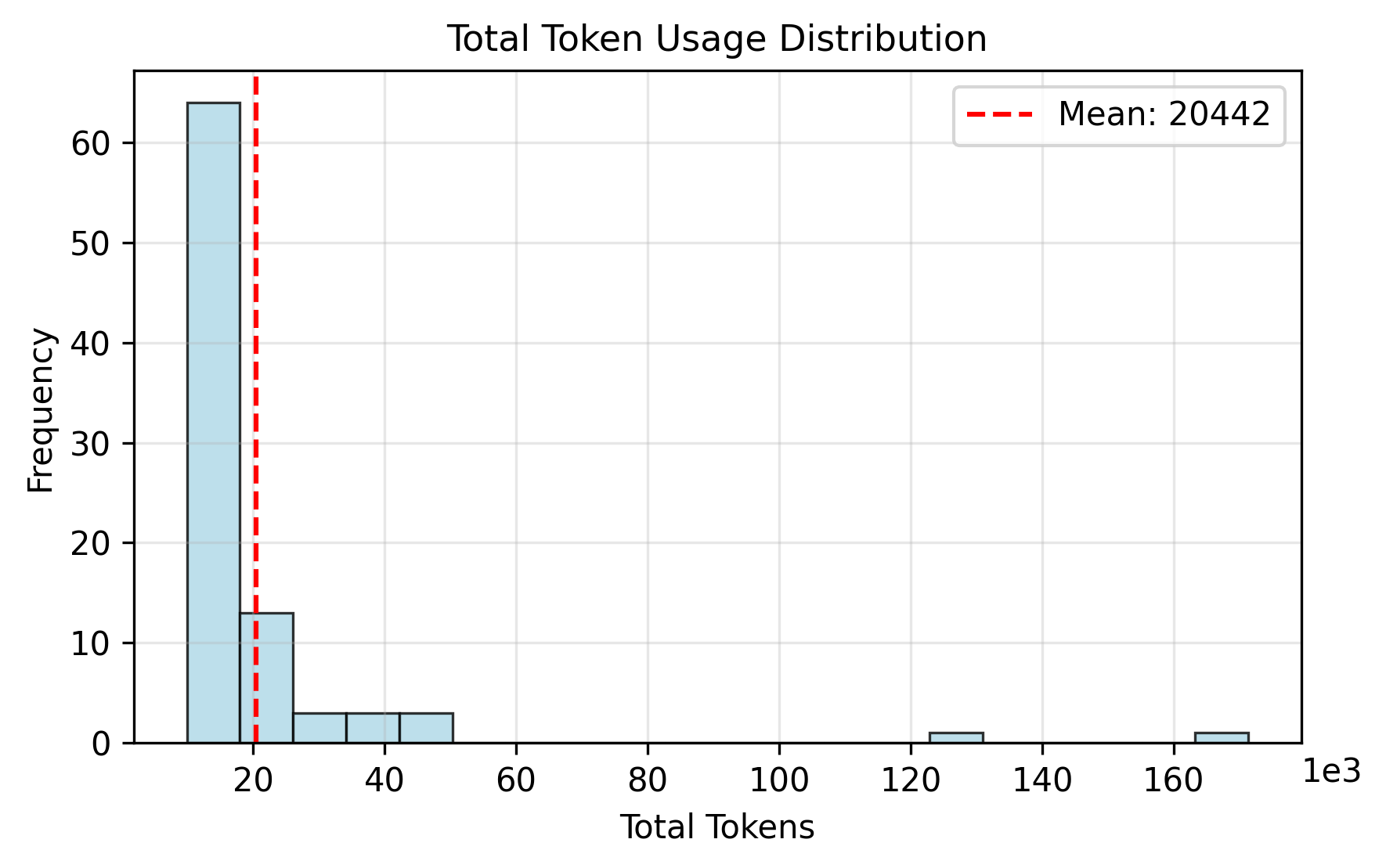}
    \caption{Distribution of number of tokens used for each question. All input, output and (context) cache tokens are counted.}
    \label{fig:token-distribution}
\end{subfigure}

\caption{Plots of the distributions of the number of turns and tokens used for each question of \textit{graph-agent-bench-ln-v1} (Sonnet-5-5).}
\label{fig:two-plots}
\end{figure}

We present some basic statistics in Figure~\ref{fig:two-plots}. 
For the most part, the model responds well to the question set.
Figure~\ref{fig:turns-distribution} shows the mean number of turns is generally low. This corresponds to the client calling basic pre-processing tools such as \texttt{get\_neo4j\_schema}, the appropriate algorithm tool, and doing some post-processing for each question. The few edge cases with more turns and tokens are path-related questions that ask for all paths or full trees from a single source (e.g., Dijkstra, Delta-Stepping, Bellman-Ford and minimum spanning tree). One example is: \textit{I would like to use the Dijkstra single-source shortest path to find all paths from Arsenal, using distance as the weight. Provide the paths with costs as answers.} The algorithm ran successfully on the first call, but its output covered all stations and exceeded the maximum tool output size. The agent therefore had to re-run the algorithm in mutate mode and recompute the paths through several paginated Cypher queries, which explains the high token usage.

 Figure~\ref{fig:token-distribution} complements Figure~\ref{fig:turns-distribution} by showing the distribution of number of tokens used for each question.

Table~\ref{tab:basic-evals} shows that, for all six models, the LLM calls the necessary algorithm tools (tool recall between 0.92 and 0.99), and for the five reasoning-capable models with mostly correct parameters (parameter match between 0.90 and 0.96). Tool precision is lower, between 0.54 and 0.83, because the models also make calls that are not expected. GPT-5.5 is best on the tool-call metrics (precision, recall, F1, parameter match). Answer match is on par with GPT-5.6-sol (0.735 vs 0.755), a difference within noise (about 2 questions out of 88). Interestingly, answer match tends to follow tool precision and F1: models that make fewer unnecessary or wrong tool calls tend to give better final answers. Tool recall and parameter match tell us much less, since nearly every model calls the expected tools with mostly correct parameters.




\begin{table}[h]
\centering
\caption{Evaluation metrics summary for six LLMs with the same tools. The first four metrics evaluate intermediate tool calls. The last metric evaluates the final output of the agent. Tool Precision = unique correct calls / total actual calls. Tool Recall = unique correct calls / expected calls. Parameter Match = correct parameter values / all parameters. Answer Match = matched items / total expected items. All metrics are macro-averaged across questions.}
\label{tab:basic-evals}
\begin{tabular}{lccccc}
\toprule
\textbf{Model} & \textbf{Tool Precision} & \textbf{Tool Recall} & \textbf{Tool F1} & \textbf{Param. Match} & \textbf{Answer Match} \\
\midrule
GPT-4o        & 0.536 & 0.923 & 0.632 & 0.725 & 0.295 \\
GPT-5.5       & 0.826 & 0.985 & 0.881 & 0.957 & 0.735 \\
GPT-5.6-sol   & 0.735 & 0.977 & 0.817 & 0.907 & 0.755 \\
Fable-5-1     & 0.619 & 0.972 & 0.741 & 0.898 & 0.696 \\
Opus-5-5      & 0.620 & 0.980 & 0.746 & 0.937 & 0.736 \\
Sonnet-5-5    & 0.613 & 0.970 & 0.737 & 0.930 & 0.658 \\
\bottomrule
\end{tabular}
\end{table}

For most of the questions, the agent finds the correct answer.
We present a few cases of ambiguity that occur by the LLM's  processing of results. In the question \textit{``Run k-core decomposition and return the core value for Stockwell station''}, the expected answer is \texttt{Stockwell:4}, which follows from projecting \texttt{LINK} as undirected. The database stores every link in both directions, so this undirected projection doubles Stockwell's degree. Four of the six models answer 4. Opus checks the links, treats the doubling as an artefact, reprojects without duplicates and answers 2. GPT-4o answers 8, because its final projection matches each stored link in both directions on top of the undirected setting. Overall, the models handle the projection differently, and this causes different and sometimes wrong answers even when the algorithm and its parameters are right.

Another source of variation is the wording of the question itself. GPT-4o answers the same question differently under two wordings. Both ask for the five stations closest to Paddington (\textit{``Which 5 stations are geographically closest to Paddington?''} and \textit{``What are the five stations located nearest to Paddington?''}), and the expected list is the five nearest by straight-line distance. The other five models answer both wordings correctly. GPT-4o gets the first wording right in the same way. On the second it reads ``nearest'' as proximity along the network, queries the \texttt{LINK} neighbours of Paddington ordered by link distance and keeps the first five rows. As a result, the rows contain duplicates and the final answer lists only three stations. The failure therefore combines a different reading of the question with a result that is not checked against the requested number of stations.

A third case is the longest path to Notting Hill Gate (\textit{``Execute the longest path algorithm to find ...''}). The algorithm requires a graph without cycles. Since every link is stored in both directions, a station can be reached from any station connected to it and back, so the network is full of cycles. Therefore, the algorithm returns that no path was found as expected. Still, although all runs pass the parameters, none matches the expected answer. Half of the runs report textually that no path exists. The other half remove links until the graph is acyclic and return some path. Models therefore differ in whether they stop at the tool's result or work around it and the workaround changes the answer.

Overall, the models reliably call the right algorithms (Table~\ref{tab:basic-evals}), while answer match varies from 0.30 to 0.76. As the three cases show, much of this gap comes from some choices around the call: how the graph is projected, how the question is read and whether a tool result is accepted or worked around.

\subsection{Results on real-world money laundering detection}
\label{sec:bybit}

The Bybit database contains around 17 million Ethereum transfers in a time window. The data is noisy and contains irrelevant activity, including look-alike addresses created by address poisoning. The agent receives one question. It is asked to identify with GDS algorithms the addresses used to launder funds stolen from a given address and to return a ranked list of up to 100 addresses. The question names the stolen-from address.

The reference is the list of 51 addresses published by the FBI \citep{fbi2025bybit}. The list is potentially partial, so the precision reported below is a lower bound. We extract the addresses and compute recall as the number of listed addresses found divided by 51, precision as the number found divided by the number of addresses returned and their F1. A run that fails or returns no list is scored zero.

For these experiments we used GPT-5. The choice comes from the rationale that this LLM's knowledge cutoff (30 September 2024) precedes the theft (February 2025), so the model cannot have seen the incident in training.  We run it 15 times at reasoning effort medium and 15 times at high, in two arms: with the GDS agent and without the GDS agent. In the second arm the model has no MCP tools and instead writes and runs Python code that can use the \texttt{neo4j} and \texttt{graphdatascience} clients and has credentials to connect to the Neo4j instance. The comparison therefore isolates the agent's tools and skill, not the GDS algorithms: 27 of the 30 runs in the second arm call GDS through the Python client. 

\begin{table}[h]
\centering
\caption{GPT-5 on Bybit, 15 runs per row. Wallets is the number of the 51 FBI addresses found per run, given as mean (median); failed runs count as zero. High and low runs are the runs that find at least 36 and at most 7 of the 51 addresses (recall of at least 0.7 and at most 0.15). Recall, precision and F1 are shown in Figure~\ref{fig:bybit_scores} and duration, tokens and turns in Figure~\ref{fig:bybit-dur-tokens}.}
\label{tab:bybit}
\begin{tabular}{llcc}
\toprule
\textbf{Arm} & \textbf{Effort} & \textbf{Wallets (of 51)} & \textbf{High / low runs} \\
\midrule
GDS agent     & medium & 27.0 (30) & 6 / 4 \\
GDS agent     & high   & 31.6 (39) & 8 / 2 \\
w/o GDS agent & medium & 18.4 (2)  & 6 / 8 \\
w/o GDS agent & high   & 19.4 (27) & 4 / 5 \\
\bottomrule
\end{tabular}
\end{table}

The runs are not uniform: many reach a recall of 0.76 to 0.82, a few collapse to a recall below 0.15. The GDS agent scored low in 6 runs, against 13 without GDS agent (3 are failed runs). Precision is a lower bound and it's low in every row (median at most 0.40) because the answers are long (62 to 100 addresses) against 51 reference addresses. Raising the effort improves the scores for the GDS agent, which suggests that it potentially works better as the LLMs get better. The failed runs are due to timeout.  Overall, with a single question the GDS agent finds a median of 30 of the 51 wallets at medium effort and 39 at high effort.

Figure~\ref{lst:bybit} shows one run at high effort. The agent handles the truncated PageRank output correctly: it reruns the algorithm in mutate mode on a graph filtered by an amount percentile and reads the full scores with Cypher. It repeats the analysis at three thresholds, combines PageRank, HITS and betweenness and uses the number of earlier transactions to set aside infrastructure such as WETH and a router. It also has weaknesses. The second method scores only the candidates of the first, so it is not an independent validation. Pairs of near-identical addresses remain in the list, because the agent never inspects the assets involved.

\begin{lstlisting}
You are a senior data scientist. You have a database of raw Ethereum transfers in a time window. The data is raw and may contain noise, misleading activity and quality issues. Funds were stolen from 0x1db92e2eebc8e0c075a02bea49a2935bcd2dfcf4. Using GDS algorithms, identify the addresses used to launder them. Return a ranked list of up to 100 addresses, most confident first. 
...


- neo4j-gds - pagerank (MCP)(graph_name: "eth_launder_h4", mode: "stream", node_identifier_property: "address", damping_factor: 0.85, relationship_weight_property: "amount", source_nodes: ["0x1db92e2eebc8e0c075a02bea49a2935bcd2dfcf4"], tolerance: 1e-06, max_iterations: 50)
  > Warning: only the first 500 of 7,466 rows are shown. They are in internal node ID order: not sorted by score, not a random sample. The 6,966 rows not shown may contain the most important results. Do not rank, estimate, generalise or pick examples from these rows. For the full result, run the algorithm in mutate mode and query the property with Cypher (gds.graph.nodeProperties.stream with WHERE, ORDER BY and LIMIT).
  >
  >      nodeId     score                                    nodeName
  ... +502 lines (ctrl+r to expand)
 
- neo4j-gds - read_neo4j_cypher (MCP)(query: "MATCH p=(s:Address {address: '0x1db92e2eebc8e0c075a02bea49a2935bcd2dfcf4'})-[:TRANSFER*1..4]->(:Address)
WHERE ALL(r IN relationships(p) WHERE r.asset='ETH')
...
RETURN percentileCont(r.amount, 0.5) AS p50,
       percentileCont(r.amount, 0.75) AS p75,
       percentileCont(r.amount, 0.9) AS p90,
       percentileCont(r.amount, 0.99) AS p99, ...")
  > [
  >   {
  >     "p50": 53.2668026515573,
  ... +8 lines (ctrl+r to expand)
 
...
 
- neo4j-gds - project_graph_cypher (MCP)(graph_name: "eth_launder_h4_p99", cypher_query: "MATCH (s:Address {address: '0x1db92e2eebc8e0c075a02bea49a2935bcd2dfcf4'})
MATCH p = (s)-[rs:TRANSFER*1..4]->(x:Address)
WHERE ALL(r IN rs WHERE r.asset = 'ETH')
...
WHERE r.asset = 'ETH' AND m IN nodes AND r.amount >= 311.85351455120934
RETURN gds.graph.project($graph_name, n, m, ...)")
  > {
  >   "graphName": "eth_launder_h4_p99",
  >   "nodeCount": 145,
  ... +4 lines (ctrl+r to expand)
 
- neo4j-gds - pagerank (MCP)(graph_name: "eth_launder_h4_p99", mode: "mutate", mutate_property: "ppr", node_identifier_property: "address", damping_factor: 0.85, relationship_weight_property: "amount", source_nodes: ["0x1db92e2eebc8e0c075a02bea49a2935bcd2dfcf4"], tolerance: 1e-06, max_iterations: 50)
  > ran_iterations=21 did_converge=True centrality_distribution={'p99': 0.1275005331262946, 'min': 0.0, 'max': 0.15000057220458982, 'p90': 0.0027093878015875816, 'mean': 0.0031163101116644925, 'p50': 0.0005709445104002953, 'p999': 0.15000057127326727, 'p95': 0.0027093878015875816, 'p75': 0.0027093878015875816} pre_processing_millis=0 compute_millis=7 post_processing_millis=3 mutate_millis=0 node_properties_written=145 configuration={'jobId': '6ce26018-19d3-4ae7-bb14-5293a0d511f5', 'maxIterations': 50, 'sourceNodes': [202], 'scaler': 'NONE', 'logProgress': True, 'nodeLabels': ['*'], 'sudo': False, 'dampingFactor': 0.85, 'relationshipTypes': ['*'], 'mutateProperty': 'ppr', 'tolerance': 1e-06, 'concurrency': 4}
 
- neo4j-gds - stream_node_properties (MCP)(graph_name: "eth_launder_h4_p99", node_properties: "ppr", node_labels: "Address", db_node_properties: ["address", "active_before", "tx_count_before"])
  >      nodeId       ppr                                     address  active_before  tx_count_before
  > 0       202  0.150000  0x1db92e2eebc8e0c075a02bea49a2935bcd2dfcf4           True               42
  > 1       473  0.127500  0x47666fab8bd0ac7003bce3f5c3585383f09486e2          False                0
  ... +143 lines (ctrl+r to expand)
 
...

- Assumptions and checks
  ...
  - Method 1 (amount-weighted Personalized PageRank, PPR): projected a value-filtered subgraph at p99 (>=311.85 ETH), ran PPR personalized to the victim (weighted by amount). This highlights high-probability downstream recipients and aggregators while suppressing dust. Most high scorers were "fresh" addresses (active_before=false).
  - Method 2 (bridge/flow structure): on a broader p90 graph (>=210.33 ETH), ran betweenness centrality to identify bridge/mixing nodes that sit on many shortest routes of the flow, then cross-checked with direct outflows of 0x4766... to enumerate the immediate 10,000 ETH splits. This method independently singled out the same pivot and mixer nodes.
  - Sensitivity: Re-ran PPR on the broader p90 graph but restricted the result readout to the Method-1 candidates. Absolute PPR scores shifted (e.g., 0x528292... fell from ~0.0053 to ~0.00034 as mass spread to more edges), but the top addresses by both PPR and betweenness remained in the leading group: 0x4766... (pivot), direct 10k recipients, and mixing funnels such as 0x528292..., 0x1fa386..., 0x344ac3..., 0xa1198f..., 0x023b29....
  - Quality controls: Excluded obvious infrastructure with extreme tx_count_before (e.g., WETH 0xc02a..., large router 0xba3c...), unless they are uniquely central in the victim's subgraph. One such hub, 0xd37bbe..., is extremely central in our subgraph; we keep it but rank it lower than the fresh laundering cluster.
  ...
 
  ANSWER
  0x47666fab8bd0ac7003bce3f5c3585383f09486e2
  0xb4a862a81abb2f952fca4c6f5510962e18c7f1a2
  0x1eb27f136bfe7947f80d6cee3cf0bfdf92b45e57
  0xb72334cb9d0b614d30c4c60e2bd12ff5ed03c305
  0xbca02b395747d62626a65016f2e64a20bd254a39
  ... (59 more addresses, 64 in total)


\end{lstlisting}
\captionof{figure}{Condensed transcript of one Bybit run. The full listing is in Figure~\ref{lst:bybit_full}.}\label{lst:bybit}

\vspace{0.5cm}

Figure~\ref{lst:bybit_failed} (Appendix~\ref{ap:bybit-db}) shows a run with very low recall (1 of 60 returned addresses is correct), in which the agent made a wrong assumption. After its GDS calls failed with memory errors, it computed two rankings in Cypher and trusted the one based on net inflow. It assumed that addresses with high onward outflow are only intermediaries and excluded them, including the address that the run in Figure~\ref{lst:bybit} ranks first. But in laundering, funds are passed on, so high outflow is expected. The ranking it discarded would have found most of the reference addresses.

\begin{figure}[ht]
\centering
\includegraphics[width=1\textwidth]{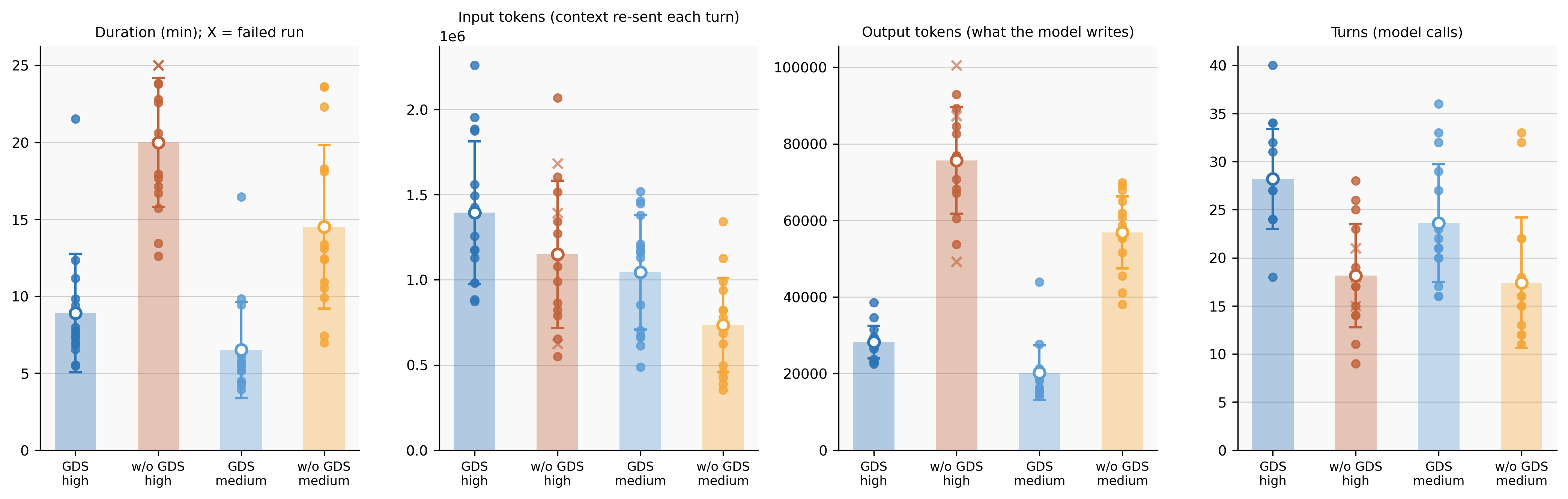}
\caption{GPT-5 on Bybit, 15 runs per arm and reasoning effort. Dots are single runs and crosses are failed runs. Bars show the mean and error bars one standard deviation. Without the GDS agent, the model writes Python using the \texttt{neo4j} and \texttt{graphdatascience} clients.}
\label{fig:bybit-dur-tokens}
\end{figure}

We have seen that the GDS agent performs better than the baseline. Now we look at additional metrics of the performance such as duration and cost. Figure~\ref{fig:bybit-dur-tokens} shows that at medium effort the median run of the GDS agent takes about 6 minutes against 13 without it, even though it needs more turns (21 against 15). The same holds at high effort. The arm without the GDS agent writes about three times more output tokens because it writes code, while the GDS agent mainly re-reads cached context, so the median run is also cheaper with the GDS agent. As a result, the GDS agent is around 2x faster with the same cost as the baseline agent.

\vspace{0.1cm}

\section{A case study of an open-ended analysis of a fly connectome \label{ss:case-flywire}}

\begin{figure}[h]
\centering
\begin{minipage}[c]{0.33\linewidth}
  \centering
  \includegraphics[width=\linewidth]{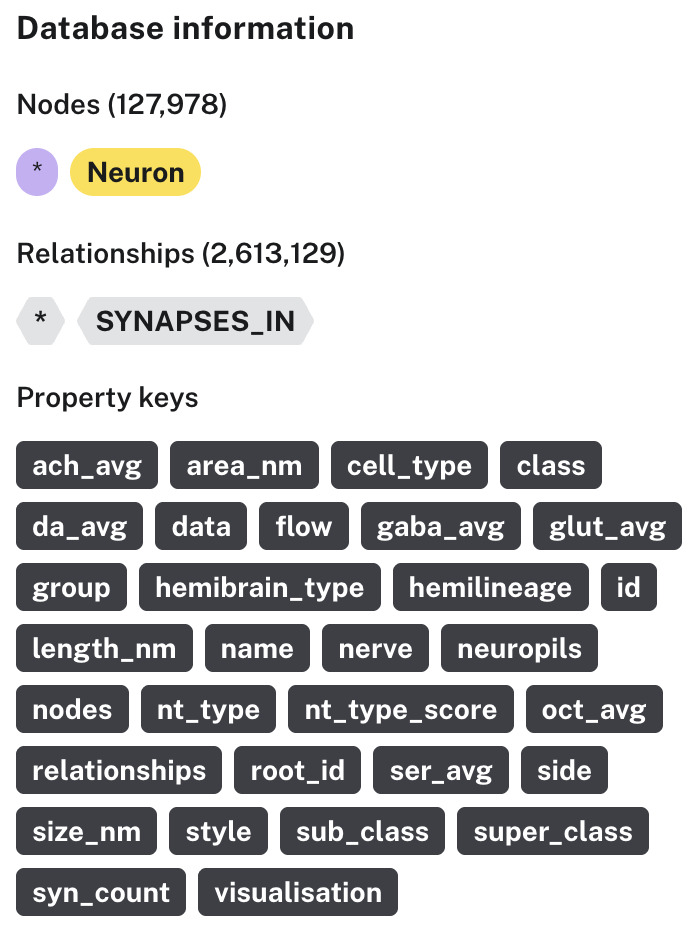}
\end{minipage}\hfill
\begin{minipage}[c]{0.63\linewidth}
  \centering
  \includegraphics[width=\linewidth]{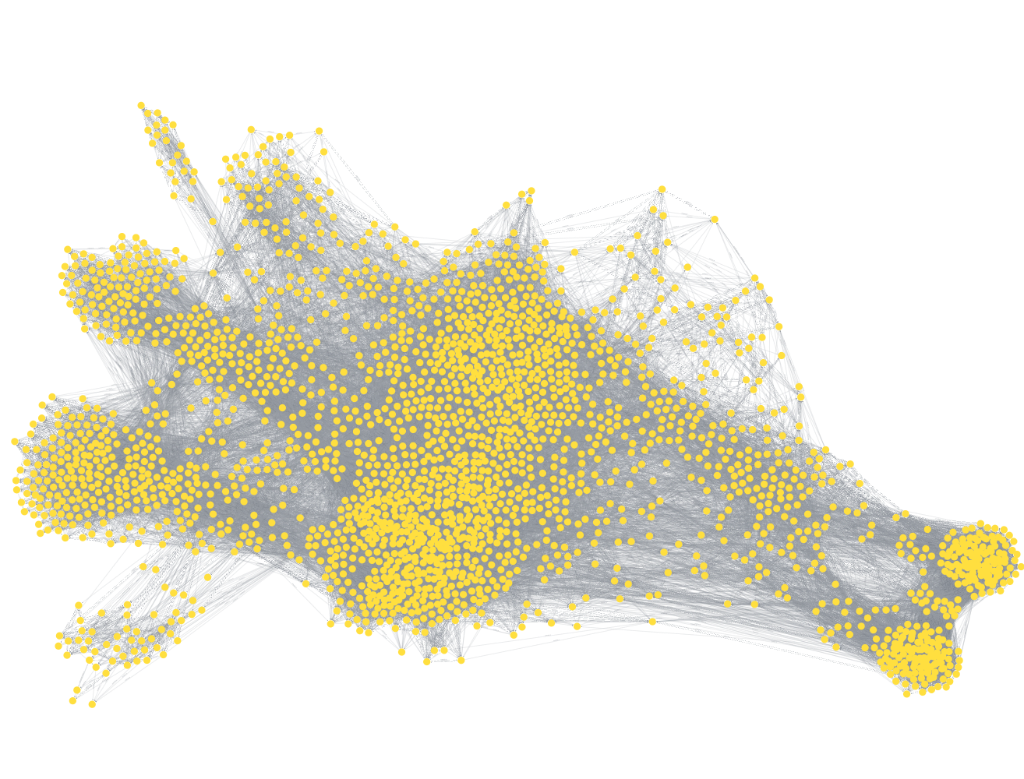}
\end{minipage}
\caption{The fly brain database. Left: statistics and schema. Right: a snapshot of a part of the graph.}
\label{fig:flywire-db}
\end{figure}

We give the agent an open-ended request on a large graph and compare its answer with a published analysis. The database holds the FlyWire connectome of the adult \textit{Drosophila} brain: 127,978 neurons and 2.6 million directed connections, with synapse counts as a relationship property (Figure~\ref{fig:flywire-db}). \citet{lin2024network} published a network analysis of this connectome. We ask the agent two open questions (Figure~\ref{lst:flywire}). The first asks it to investigate how connected the graph is as a whole and to compare its findings with random networks. The follow up asks which neurons a signal that wanders along the connections would spend most of its time on, and whether they share any biology. The first question asks for random networks, but not which kind or how to build them. Neither question names an algorithm or a statistic.

The agent works like an analyst. In the first question it measures the connected components, clustering and community structure of the brain and compares each with random networks; in the second it ranks the neurons and profiles the top ones. Four of its choices matter. First, the random networks keep every neuron's incoming and outgoing connection counts and are measured exactly like the brain; when the first construction exceeded the database's memory limit, the agent rewrote the query instead of shrinking the comparison. Second, the projections are chosen accurately: the directed projection keeps the neurons without connections and the agent notices that reciprocal connections appear twice in the undirected projection and rebuilds it with each connected pair once without doubling. Third, it checks convergence: a PageRank run with a tolerance of $10^{-8}$ stopped at 100 iterations without converging and the agent reran it with a looser tolerance and more iterations. Fourth, it does not rank on a truncated result: when the streaming tool warned that it had returned only 500 of 127,978 rows, the agent wrote the scores into the projection and read the top ranks with a Cypher query.

The agent's analysis follows the published one and its numbers are close. Its average distance of 3.9 steps, estimated from closeness centrality, matches the 3.91 of \citet{lin2024network}, and its giant components cover 97.5\% (weakly connected) and 90.4\% (strongly connected) of the neurons, against the published 98.8\% and 93.3\%. The small differences are expected: the published analysis uses slightly different graphs in different parts (\citealp{lin2024network}; code: https://github.com/murthylab/flywire-network-analysis) which do not fully match the database we have. The comparison with random networks leads to the same picture as the paper's small-world result: the brain is far more clustered (0.157 against 0.008) and more modular (0.68 against 0.12) than chance, with only slightly longer routes (3.9 against 3.15).

\begin{lstlisting}
> You have a brain wiring database in Neo4j. Using the GDS agent, investigate how connected this graph is as a whole. Make a table of the most important, well-tested findings (no more than 5), each with one number. For each finding, also give the typical value and range across several random networks to find out whether the brain is surprising compared to them. State what kind of random networks you used.

- I'll start by loading the graph tools and checking what the database contains.
- neo4j-gds - get_neo4j_schema (MCP)()
  > { "Neuron": { "type": "node", ... +189 lines (ctrl+r to expand)
- I'm projecting the brain network into memory, both with directions kept and as an undirected one.
- neo4j-gds - project_graph_cypher (MCP)(graph_name: "brain_dir", cypher_query: "MATCH (n:Neuron)
  OPTIONAL MATCH (n)-[r:SYNAPSES_IN]->(m:Neuron) RETURN gds.graph.project($graph_name, n, m, ...)")
  > { "graphName": "brain_dir", "nodeCount": 127978, ... +4 lines (ctrl+r to expand)
- neo4j-gds - weakly_connected_components (MCP)(graph_name: "brain_und", mode: "mutate", ...)
  > component_count=3122 component_distribution={... 'max': 124778, ...} ...
- neo4j-gds - strongly_connected_components (MCP)(graph_name: "brain_dir", mode: "mutate", ...)
  > component_count=12247 component_distribution={... 'max': 115661, ...} ...
- neo4j-gds - local_clustering_coefficient (MCP)(graph_name: "brain_und", mode: "mutate", ...)
  > ... average_clustering_coefficient=0.156877031188439 ...
- neo4j-gds - leiden (MCP)(graph_name: "brain_und", mode: "mutate", ...)
  > community_count=3133 ... modularity=0.6916444206987016 ...
- neo4j-gds - closeness_centrality (MCP)(graph_name: "brain_und", mode: "mutate", ...)
  > ... centrality_distribution={... 'mean': 0.245523498042922, ...} ...
- Reciprocal links create duplicate edges in the undirected version, so I'm building a cleaned
  version (each connected pair counted once) for a fair comparison with the random networks.
- neo4j-gds - project_graph_cypher (MCP)(graph_name: "brain_s", cypher_query: "... WHERE id(n) <
  id(m) OR NOT (m)-[:SYNAPSES_IN]->(n) ...")
  ... (the same measures on brain_s; Leiden repeated twice: modularity 0.681, 0.679, 0.679)
- Brain numbers are in. Now I'm building the first random network, which keeps every neuron's
  number of incoming and outgoing links but shuffles who connects to whom.
- neo4j-gds - project_graph_cypher (MCP)(graph_name: "rnd_u1", cypher_query: "MATCH
  (a:Neuron)-[:SYNAPSES_IN]->(b:Neuron) WITH collect(id(a)) AS S, collect(id(b)) AS T0 ...")
  > Error: {neo4j_code: Neo.TransientError.General.MemoryPoolOutOfMemoryError} {message: The
  allocation of an extra 142.8 MiB would use more than the limit 11.2 GiB. ...}
- Too much memory in that version, so I'm retrying with a leaner query.
- neo4j-gds - project_graph_cypher (MCP)(graph_name: "rnd_u1", cypher_query: "CALL { MATCH
  (a:Neuron)-[:SYNAPSES_IN]->() RETURN collect(id(a)) AS S } CALL { MATCH ()-[:SYNAPSES_IN]->
  (b:Neuron) WITH id(b) AS t, rand() AS r ORDER BY r RETURN collect(t) AS T } ...")
  > { "graphName": "rnd_u1", "nodeCount": 124891, ... +4 lines (ctrl+r to expand)
- neo4j-gds - local_clustering_coefficient (MCP)(graph_name: "rnd_u1", mode: "mutate", ...)
  > ... average_clustering_coefficient=0.008170259429587719 ...
- neo4j-gds - leiden (MCP)(graph_name: "rnd_u1", mode: "mutate", ...)
  > community_count=42 ... modularity=0.12272997113141473 ...
  ... (the same measures on four more random networks, and on five directed random networks for
  the strongly connected components; each network is dropped after it is measured)
  
- The brain is as connected as chance, yet far more clustered and grouped, with slightly longer routes.

  | Finding | Brain | Random typical (range) |
  |---|---|---|
  | Neurons in one linked network | 97.5% | 97.6% (identical) |
  | Neurons reaching each other both ways | 90.4% | 90.7% (identical) |
  | Average steps between neurons | 3.9 | 3.15 (3.145-3.147) |
  | Neighbours that link to each other | 0.157 | 0.008 (0.0079-0.0080) |
  | Grouping into tight communities | 0.68 | 0.12 (0.120-0.123) |

  Random networks: five shuffles keeping each neuron's incoming and outgoing link counts, with
  partners rewired. 3,087 neurons have no links at all.

> You found that this brain is nearly one connected piece. If a signal wandered through it
  following connections, which neurons would it spend most of its time on? Do those neurons have
  anything in common biologically?

- neo4j-gds - pagerank (MCP)(graph_name: "brain_dir", mode: "mutate", mutate_property: "pr_fresh",
  damping_factor: 0.85, max_iterations: 100, tolerance: 1e-08)
  > ran_iterations=100 did_converge=False ...
- neo4j-gds - pagerank (MCP)(graph_name: "brain_dir", mode: "mutate", mutate_property: "pr_conv",
  damping_factor: 0.85, max_iterations: 1000, tolerance: 1e-06)
  > ran_iterations=75 did_converge=True ...
- neo4j-gds - stream_node_properties (MCP)(graph_name: "brain_dir", node_properties: ["pr_conv"])
  > Warning: only the first 500 of 127,978 rows are shown. ... Do not rank, estimate, generalise
  or pick examples from these rows. ...
- neo4j-gds - read_neo4j_cypher (MCP)(query: "CALL gds.graph.nodeProperties.stream('brain_dir',
  ['pr_conv']) YIELD nodeId, propertyValue ... ORDER BY pr DESC LIMIT 40 RETURN ...")
  > [ { "id": 720575940621675174, ... +479 lines (ctrl+r to expand)
  ... (a cross-check with degree centrality on brain_s: top-100 overlap)

- A wandering signal would spend most of its time on a small set of hub neurons, spread across
  vision, memory and movement-command regions.

  They do share some biology:

  - Size: the top 100 have about 20 times the average wiring length.
  - Inhibition: 41% are inhibitory (GABA) neurons, versus 12% overall.
  - Role: none are sensory input cells.
  - Cross-check: a connection-count method agreed on size and inhibition, though only 43 top
    neurons overlapped.

  To read the scores back I ran one query that called a GDS procedure.
\end{lstlisting}
\captionof{figure}{The agent does graph data analysis by measuring connectivity, clustering, community structure, runs PageRank and compares each with correctly defined random networks.}
\label{lst:flywire}

Here we show that from a general request the agent reproduces the main statistics of a published analysis of a real-world graph and that it chooses the random networks, projections and checks that this analysis depends on. Its results cover some of the findings of the published paper. Such an analysis would otherwise need a researcher who knows which graph measures to use and how to compute them at this scale. The agent can therefore act as a helpful assistant for the first steps of an exploratory analysis.

\section{Conclusion}

In this technical report, we introduced the GDS agent, an MCP server that exposes graph algorithms as tools, together with a skill that guides the workflow and an LLM as the MCP client. Users can collaborate with the agent in plain language on a wide range of graph tasks that are hard to solve with a query language alone or with LLMs that reason over textualised graphs. This extends the usefulness of LLMs to large-scale private and enterprise knowledge graphs and lowers the barrier to using graph analytics libraries.

Our benchmarks show that the agent picks the right tools, mostly sets their parameters correctly and answers many non-trivial graph questions correctly, from centrality and path-finding to community detection. The graph stays in the database and is never given to the model as text, so the agent is not limited by the context window and works on graphs far larger than those used by most existing benchmarks. It also handles practical difficulties: it chooses projections, notices when a tool output is cut and recovers the full result, checks whether an algorithm result is as expected, does statistical analysis.

The open-ended experiments show that this carries over to real analyses. With a single question over millions of transfers, the agent finds many of the addresses used to launder the funds of the 2025 Bybit theft and it does so fast and relatively cheaply. On the FlyWire connectome, a general request leads it to run an analysis similar to a published study and its results are close to the published ones. In both cases a user needs no graph data science expertise and can inspect the tool calls and raw data behind the answer.

In the future, we plan to add new tools to the server, reduce token usage and improve the accuracy and efficiency of tool use. We also plan to extend the benchmarks with open-ended questions and evaluate the performance of the GDS agent with external tools (e.g. coding). We expect that many of these findings apply to other graph agents built on any graph database and graph data science library and to reasoning agents more generally.

\begin{ack}
The authors thank Jon Besga, Nathalie Charbel, Will Tai and the Neo4j GDS and GenAI team for contributions to developing the prototype and helpful discussions.
\end{ack}

\clearpage
\bibliographystyle{plainnat}
\bibliography{references}


\appendix

\clearpage
\section{List of tools in GDS agent}\label{ap:tool-list}
We provide the complete list of the 69 tools of the server (version 1.2.0): 66 tools that are always available, and 3 tools that are only available in session mode.
Schema, projection and result tools:
\begin{lstlisting}
get_neo4j_schema, project_graph_cypher, list_graphs, get_graph_info, drop_graph, stream_node_properties, stream_relationship_properties, stream_relationships, read_neo4j_cypher
\end{lstlisting}
Centrality algorithm tools:
\begin{lstlisting}
article_rank, articulation_points, betweenness_centrality, bridges, CELF, closeness_centrality, degree_centrality, eigenvector_centrality, harmonic_centrality, HITS, pagerank
\end{lstlisting}
Community algorithm tools:
\begin{lstlisting}
approximate_maximum_k_cut, conductance, HDBSCAN, k_1_coloring, k_core_decomposition, k_means_clustering, label_propagation, leiden, local_clustering_coefficient, louvain, modularity_metric, modularity_optimization, speaker_listener_label_propagation, strongly_connected_components, triangle_count, weakly_connected_components
\end{lstlisting}
Similarity algorithm tools:
\begin{lstlisting}
k_nearest_neighbors, node_similarity
\end{lstlisting}
Path finding algorithm tools:
\begin{lstlisting}
a_star_shortest_path, all_pairs_shortest_paths, bellman_ford_single_source_shortest_path, breadth_first_search, delta_stepping_shortest_path, depth_first_search, dijkstra_single_source_shortest_path, find_shortest_path, longest_path, max_flow, minimum_directed_steiner_tree, minimum_weight_spanning_tree, prize_collecting_steiner_tree, random_walk, yens_shortest_paths
\end{lstlisting}
Node embedding tools:
\begin{lstlisting}
fast_rp, graph_sage_predict, graph_sage_train, hashgnn, node2vec
\end{lstlisting}
Machine learning tools:
\begin{lstlisting}
drop_model, list_models, predict_link_prediction, predict_node_classification, predict_node_regression, train_link_prediction_model, train_node_classification_model, train_node_regression_model
\end{lstlisting}
Session tools (session mode only):
\begin{lstlisting}
list_sessions, create_session, delete_session
\end{lstlisting}

\section{Example tool specification - Yen's algorithm}\label{ap:yens}
\begin{lstlisting}
types.Tool(
    name="yens_shortest_paths",
    description="Compute the k shortest paths between source_node and target_node using Yen's algorithm (non-negative weights).",
    inputSchema={
        "type": "object",
        "properties": {
            "graph_name": {"type": "string", "description": "Name of an in-memory projected graph; inspect with list_graphs or create with project_graph_cypher."},
            "source_node": {"description": "Identifier of the source node."},
            "target_node": {"description": "Identifier of the target node."},
            "node_identifier_property": {"type": "string", "description": "Node property used for readable identifiers; inspect with get_neo4j_schema."},
            "k": {"type": "integer", "description": "Number of shortest paths to compute."},
            "relationship_weight_property": {"type": "string", "description": "Relationship property to use as weights (unweighted if omitted)."},
            "mode": {"type": "string", "enum": ["stream", "mutate"], "description": "Execution mode (default: 'stream')."},
            "mutate_relationship_type": {"type": "string", "description": "Relationship type for new relationships written to the projected graph (required when mode is 'mutate')."},
        },
        "required": ["graph_name", "source_node", "target_node", "node_identifier_property"],
    },
)
\end{lstlisting}

\section{graph-agent-bench-ln-v1} 
Detailed results for  \textit{graph-agent-bench-ln-v1}.

\begin{figure}[h]
  \centering
  \includegraphics[width=0.98\textwidth]{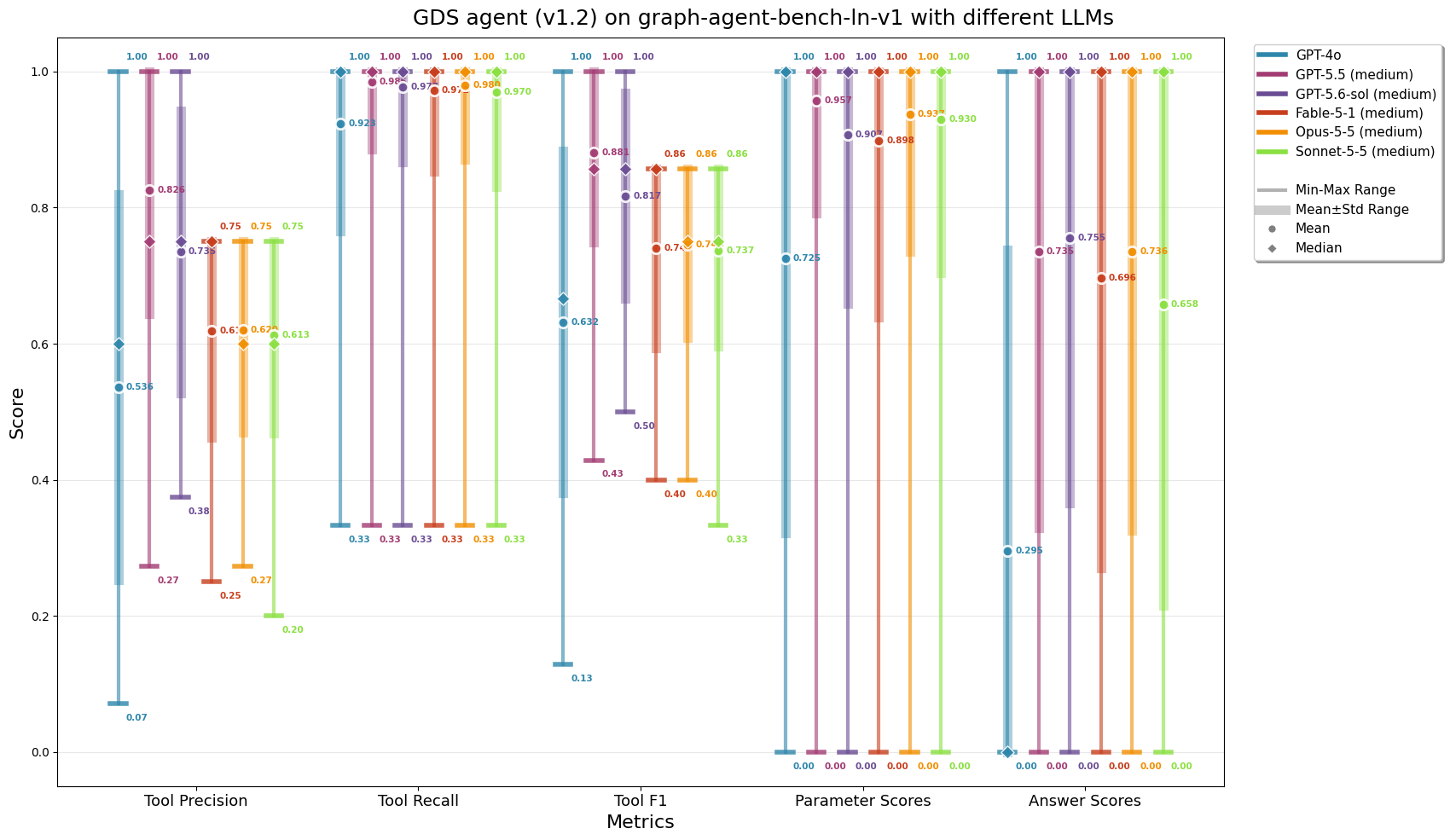}
  \caption{Different LLMs with the same set of tools benchmarked on graph-agent-bench-ln-v1. We can observe that the models differ in how well they use the tools.}
  \label{fig:ln_llm_model_comparison}
\end{figure}

Figure~\ref{lst:farthest} shows an example question on the London map: which two stations are furthest apart. The agent projects the network as an undirected graph and runs all-pairs shortest paths. The agent checks that the network is connected and states its assumption that distance counts hops, ignoring line changes and travel time.

\begin{lstlisting}
```
> You are a graph data science assistant that helps analyse the graph in my database. I have a transport network. Which two stations are furthest apart, and how many stops is the journey? Provide an accurate explanation

- I'll start by loading the graph tools and checking the schema.
- ToolSearch(query: "select:mcp__remote-devices__neo4j-gds__get_neo4j_schema,mcp__remote-devices__neo4j-gds__list_graphs,mcp__remote-devices__neo4j-gds__read_neo4j_cypher,mcp__remote-devices__neo4j-gds__all_pairs_shortest_paths,mcp__remote-devices__neo4j-gds__project_graph_cypher,mcp__remote-devices__neo4j-gds__find_shortest_path,mcp__remote-devices__neo4j-gds__get_graph_info", max_results: 10)
  > <functions>
  > <function>{"description": "Returns nodes, their properties (with types), and relationship types using native Neo4j schema introspection.", "name": "mcp__remote-devices__neo4j-gds__get_neo4j_schema", "parameters": {"properties": {}, "type": "object"}}</function>
  ... +7 lines (ctrl+r to expand)
- neo4j-gds - get_neo4j_schema (MCP)()
  > {
  >   "UndergroundStation": {
  >     "type": "node",
  ... +95 lines (ctrl+r to expand)
- neo4j-gds - list_graphs (MCP)()
  > {
  >   "graphs": [],
  >   "count": 0
  ... +1 line (ctrl+r to expand)
- neo4j-gds - project_graph_cypher (MCP)(graph_name: "tube", cypher_query: "MATCH (n:UndergroundStation)-[r:LINK]->(m:UndergroundStation)
RETURN gds.graph.project($graph_name, n, m, {sourceNodeLabels: labels(n), targetNodeLabels: labels(m), relationshipType: type(r)}, {undirectedRelationshipTypes: ['LINK']})")
  > {
  >   "graphName": "tube",
  >   "nodeCount": 302,
  ... +4 lines (ctrl+r to expand)
- neo4j-gds - all_pairs_shortest_paths (MCP)(graph_name: "tube")
  > Warning: this output was cut at 100,000 characters. 61,949,262 characters (100%) are missing, so the text below is incomplete. Do not count, list, summarise or pick examples from it as if it were the full result. For the full result, run the algorithm in mutate mode and query the result with Cypher.
  > {
  >   "found": true,
  ... +1030 lines (ctrl+r to expand)
- The full output was truncated, so I'll run the same algorithm directly in Cypher and return only the largest distances.
- neo4j-gds - read_neo4j_cypher (MCP)(query: "CALL gds.allShortestPaths.stream('tube') YIELD sourceNodeId, targetNodeId, distance
WHERE distance < 1.0E9 AND sourceNodeId < targetNodeId
WITH sourceNodeId, targetNodeId, distance
ORDER BY distance DESC LIMIT 5
RETURN gds.util.asNode(sourceNodeId).name AS a, gds.util.asNode(targetNodeId).name AS b, distance")
  > [
  >   {
  >     "a": "Heathrow Terminal 4",
  ... +24 lines (ctrl+r to expand)
- neo4j-gds - read_neo4j_cypher (MCP)(query: "CALL gds.wcc.stream('tube') YIELD componentId RETURN count(DISTINCT componentId) AS components")
  > [
  >   {
  >     "components": 1
  ... +2 lines (ctrl+r to expand)
  
- The furthest-apart stations are **Upminster** and **Heathrow Terminal 4**, a journey of **38 hops**, which is 37 intermediate stops. Upminster is also 38 hops from **Heathrow Terminals 1, 2 & 3**, so those two pairs tie.

  - **Method:** I projected all 302 stations and their links as an undirected graph, then ran all-pairs shortest paths unweighted, so distance is the minimum number of hops between stations. The network is a single connected component, so every pair is reachable and no distant pair was left out.
  - **Interchanges:** The 38 is the fewest hops possible, and it ignores line changes and travel time. It assumes you can change lines freely at any interchange.
  - **Next longest:** Upminster to West Ruislip is 37 hops, and Upminster Bridge to either Heathrow terminal is also 37.
```
\end{lstlisting}
\captionof{figure}{Case study of finding the two stations that are furthest apart. The agent projects the network, runs all-pairs shortest paths and reruns the algorithm through Cypher when the tool output is truncated.}
\label{lst:farthest}

\section{graph-agent-bench-got-v1} \label{ap:got}
\begin{figure}[h]
  \centering
  \includegraphics[width=0.98\textwidth]{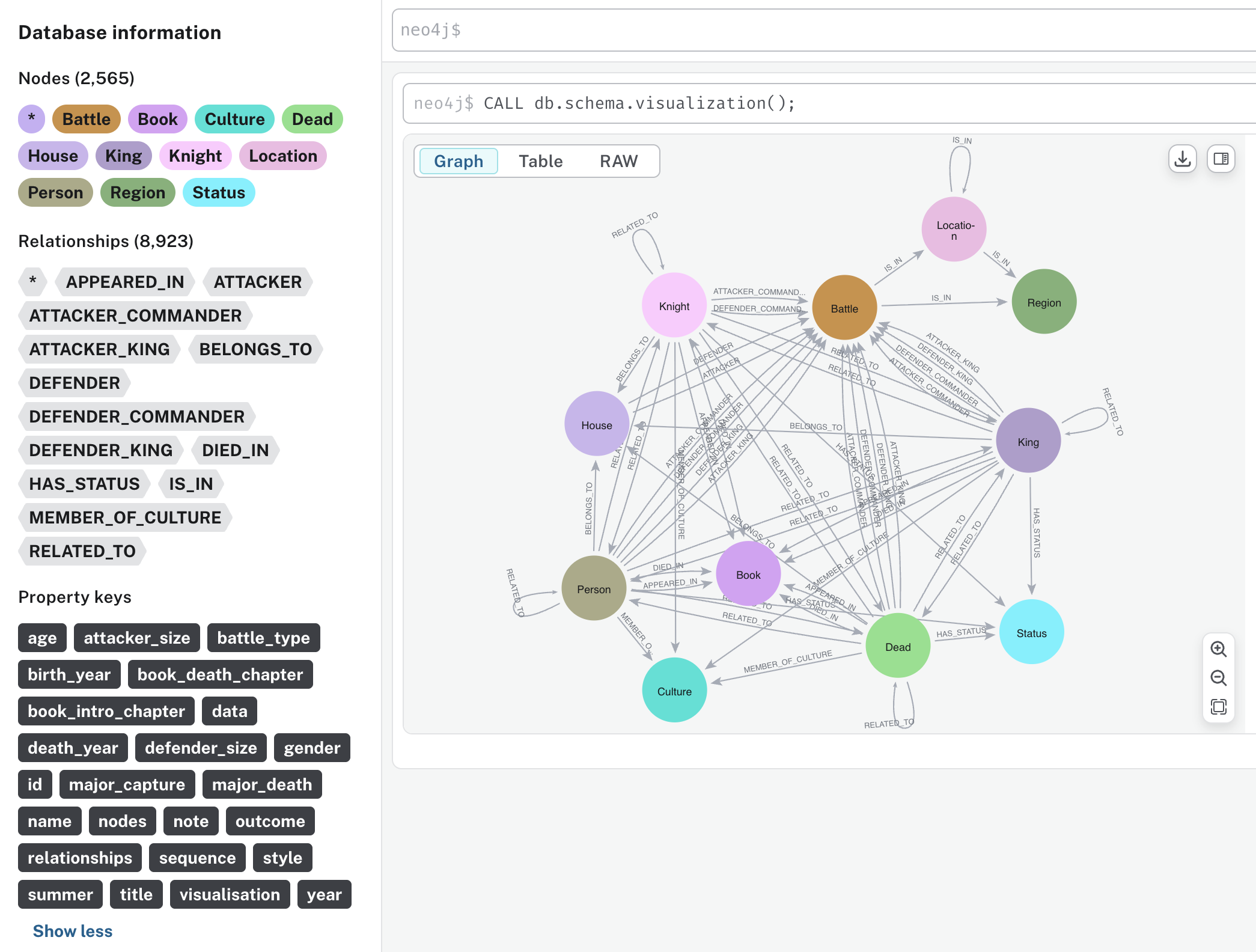}
  \caption{The dataset statistics and schema of the GoT dataset.}
  \label{fig:got_dataset}
\end{figure}

The Game of Thrones (GoT) dataset contains 2565 nodes and 8923 relationships with different nodes and relationship types as well as properties, as shown in Figure~\ref{fig:got_dataset}.

The questions follow a similar format to \textit{graph-agent-bench-ln-v1}. A few example questions are listed below.

\begin{lstlisting}
Question: Which house has the most Knights? Return the name of the house and nothing else. Return all the top house names if there are ties.
Expected Tool Calls: ["mcp__gds-agent__get_neo4j_schema", "mcp__gds-agent__project_graph_cypher", "mcp__gds-agent__degree_centrality"]
Expected Tool Parameters: {"degree_centrality": {}}
Expected Answers: Lannister

Question: If battles and the people who fought in them are organized into three interconnected communities, are Moat Cailin and the Shield Islands battles placed in the same group? Answer Yes or No and nothing else.
Expected Tool Calls: ["mcp__gds-agent__get_neo4j_schema", "mcp__gds-agent__project_graph_cypher", "mcp__gds-agent__approximate_maximum_k_cut"]
Expected Tool Parameters: {"approximate_maximum_k_cut": {"node_identifier_property": "name", "k": 3}}
Expected Answers: No
\end{lstlisting}

\begin{figure}[h]
  \centering
  \includegraphics[width=0.98\textwidth]{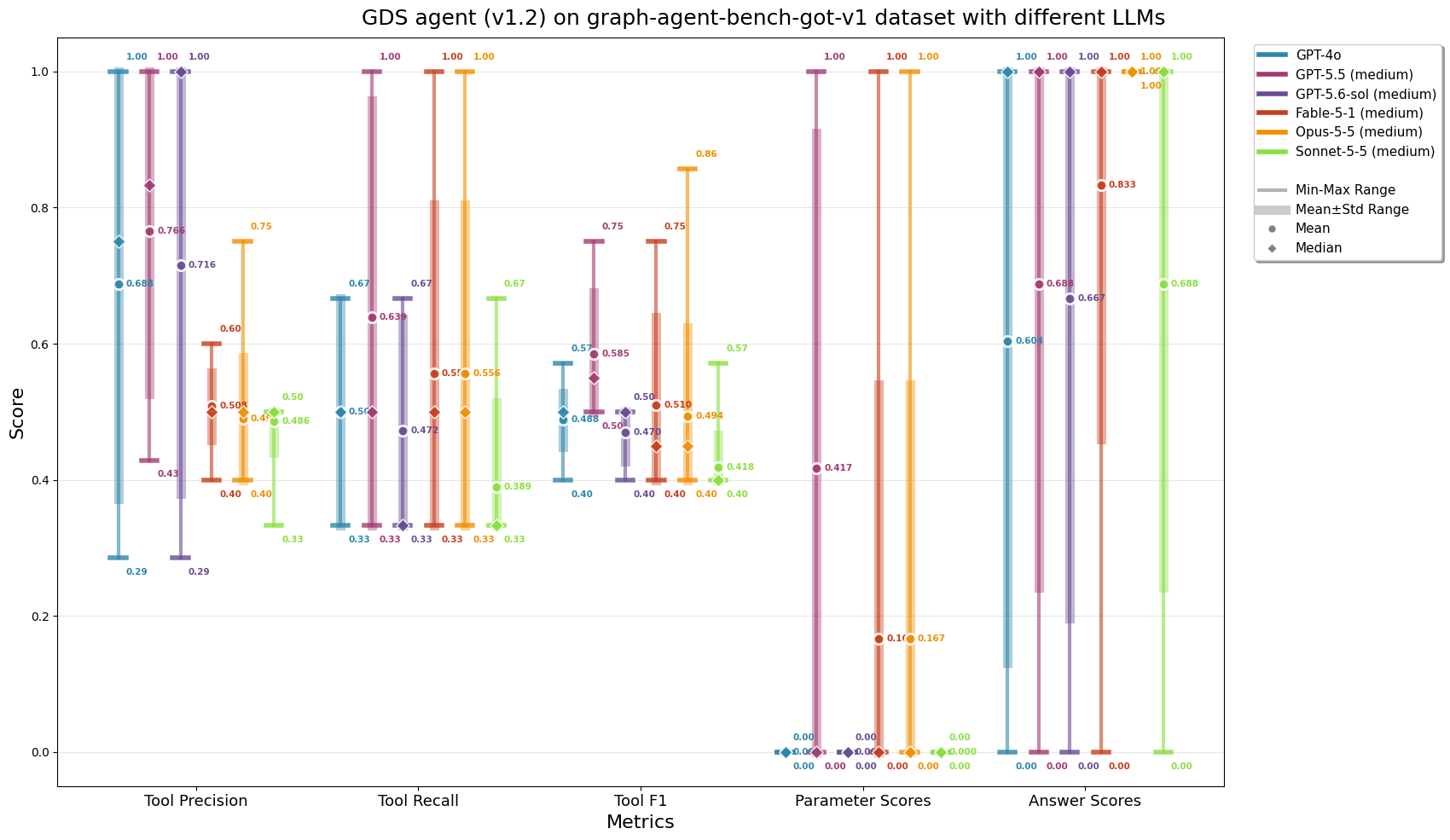}
  \caption{Benchmark results for the GDS agent on \textit{graph-agent-bench-got-v1} with different LLMs.}
  \label{fig:got_benchmark}
\end{figure}

Benchmarking results for the agent with different LLM backbones on \textit{graph-agent-bench-got-v1} are shown in Figure ~\ref{fig:got_benchmark}.

The heterogeneity of the graph poses significantly more challenges to the agent compared to the LN underground dataset, as shown by the lower scores on most metrics (Figure~\ref{fig:got_benchmark}). Opus-5-5 and Fable-5-1 have the highest answer match (1.00 and 0.83), although GPT-5.5 has the highest tool precision, recall and parameter score.

\section{Bybit database}\label{ap:bybit-db}

\begin{figure}[h]
\centering
\begin{minipage}[c]{0.33\linewidth}
  \centering
  \includegraphics[width=\linewidth]{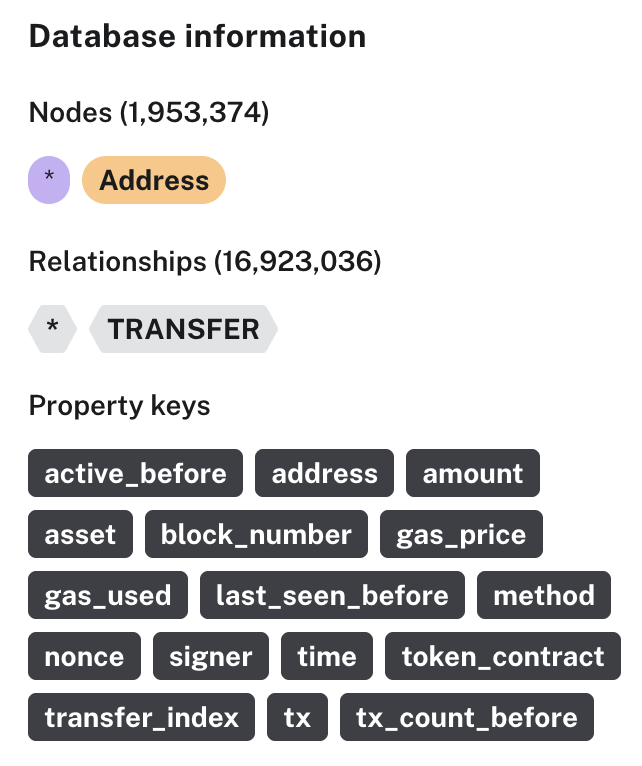}
\end{minipage}\hfill
\begin{minipage}[c]{0.63\linewidth}
  \centering
  \includegraphics[width=\linewidth]{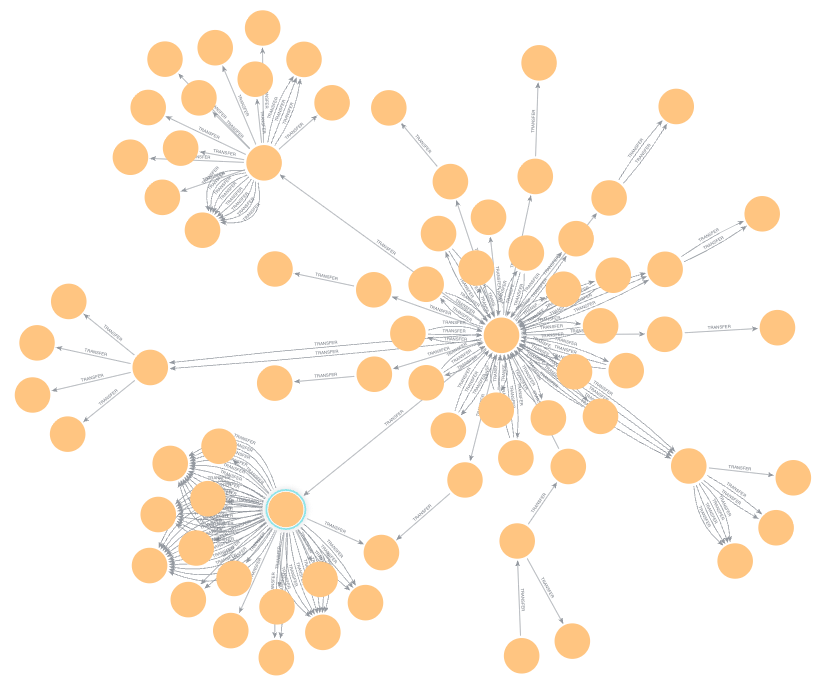}
\end{minipage}
\caption{The Bybit database. Left: statistics and schema. Right: a snapshot of the graph.}
\label{fig:bybit-db}
\end{figure}

The schema of the Bybit database is shown in Figure\ref{fig:bybit-db}.

\begin{figure}[H]
\centering
\begin{minipage}[c]{0.49\linewidth}
  \centering
  \includegraphics[width=\linewidth]{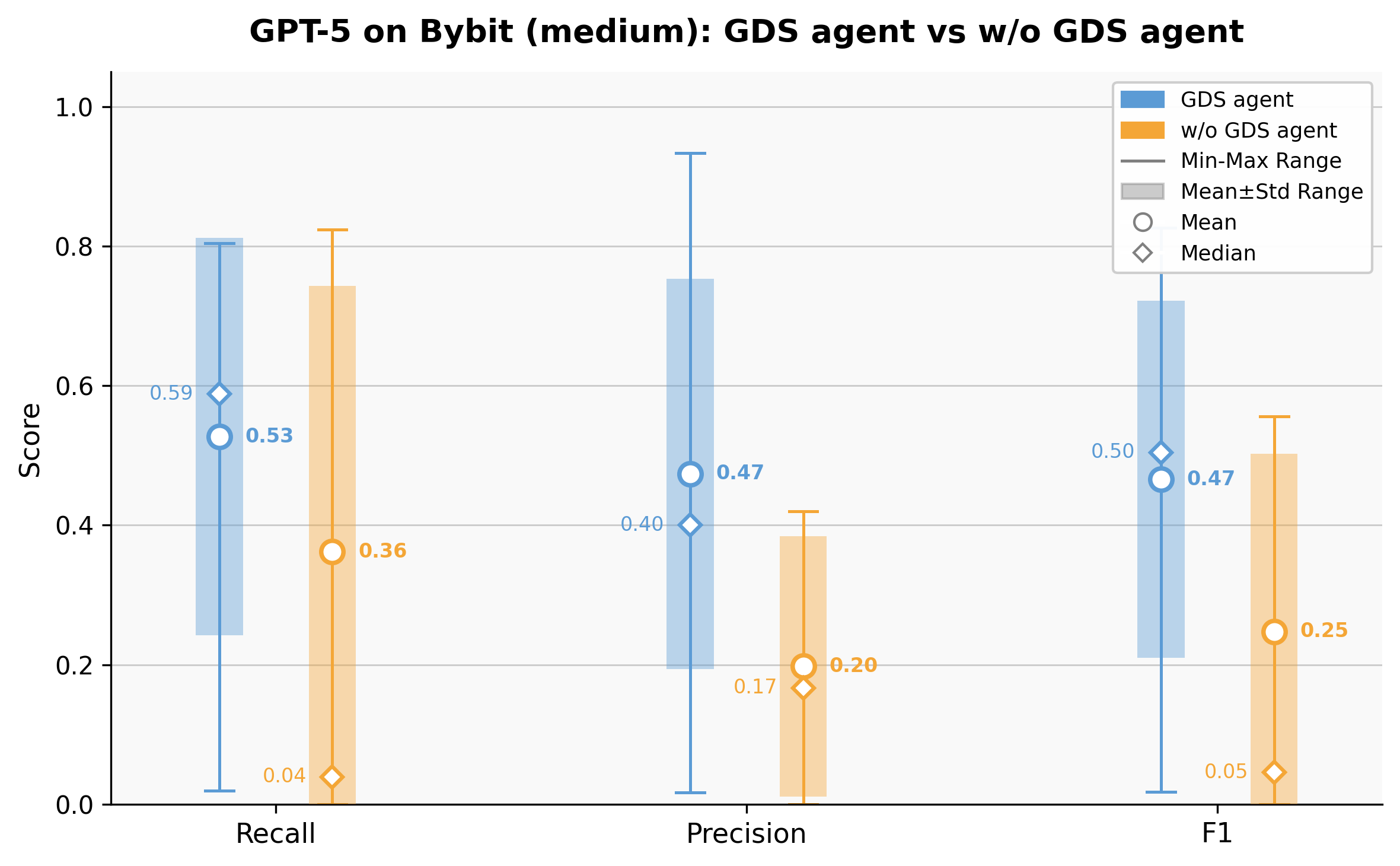}
\end{minipage}\hfill
\begin{minipage}[c]{0.49\linewidth}
  \centering
  \includegraphics[width=\linewidth]{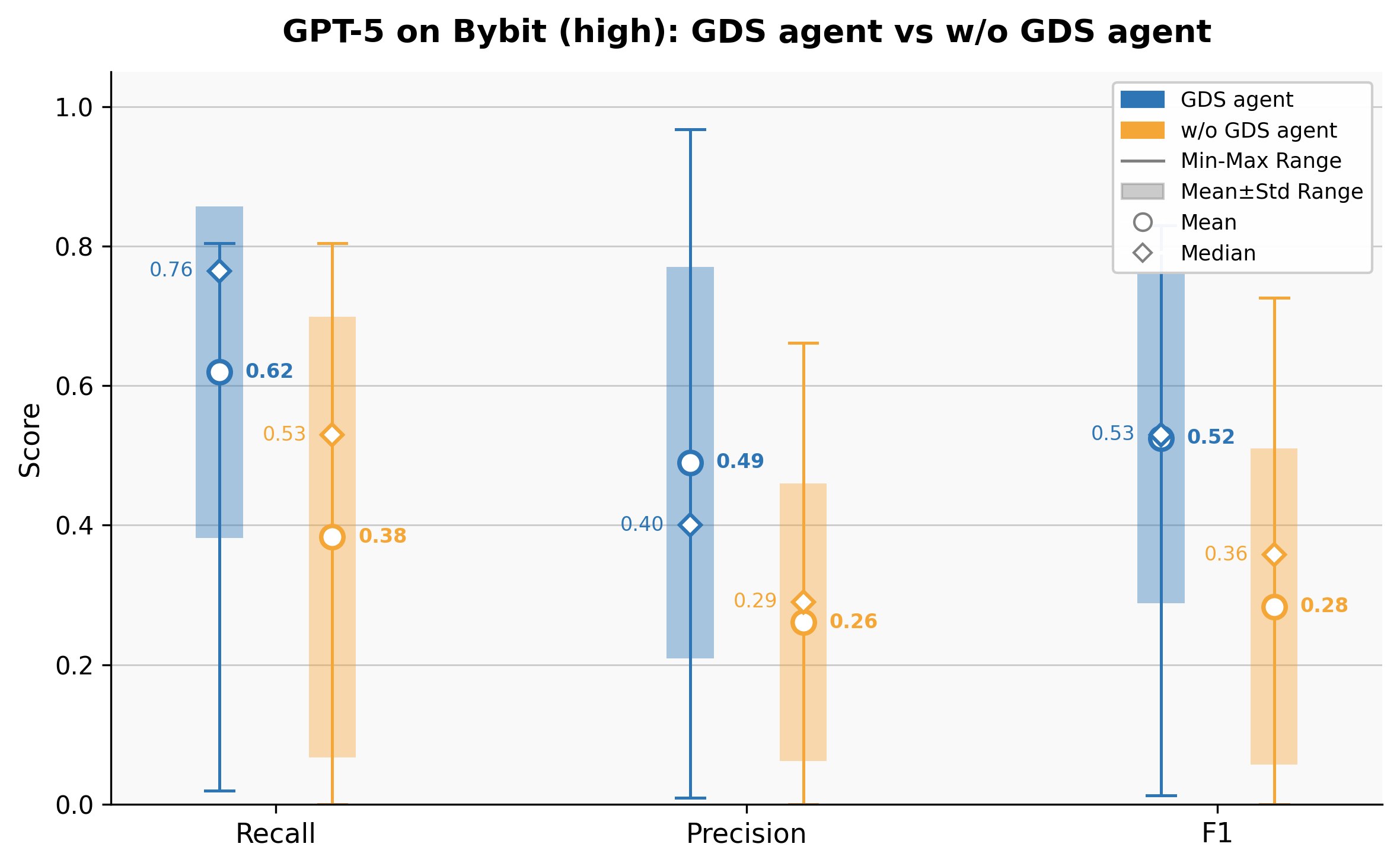}
\end{minipage}
\caption{Recall, precision and F1 of GPT-5 on Bybit, with and without the GDS agent, at medium (left) and high (right) reasoning effort. Circles are means, diamonds are medians, shaded bars show one standard deviation and whiskers the minimum and maximum over 15 runs.}
\label{fig:bybit_scores}
\end{figure}

Figure~\ref{fig:bybit_scores} shows recall, precision and F1 for both arms. The GDS agent has higher means at both efforts, mostly because the arm without it has many runs that find almost no addresses. The spread is large in both arms.

The full transcript of the run in Section~\ref{sec:bybit}.

\begin{lstlisting}
> You are a senior data scientist. You have a database of raw Ethereum transfers in a time window. The data is raw and may contain noise, misleading activity and quality issues. Funds were stolen from 0x1db92e2eebc8e0c075a02bea49a2935bcd2dfcf4. Using GDS algorithms, identify the addresses used to launder them. Do not stop at your first result. Produce your list in two different ways and compare them, test how much it changes when you change your filters or thresholds, and inspect the addresses you are least sure about. If your variants disagree, say which you trust and why, and build your answer from that one. Wrong answers and missed real addresses both count against you. Return a ranked list of up to 100 addresses, most confident first. Include only addresses you can justify; fewer is fine. Answer format: first a short summary of your assumptions, the checks you ran, and what you changed because of them. Then the heading ANSWER, followed by the addresses, one per line. Write nothing after the list.

- neo4j-gds - get_neo4j_schema (MCP)()
  > {
  >   "Address": {
  >     "type": "node",
  ... +117 lines (ctrl+r to expand)

- neo4j-gds - read_neo4j_cypher (MCP)(query: "MATCH (s:Address {address: '0x1db92e2eebc8e0c075a02bea49a2935bcd2dfcf4'})
OPTIONAL MATCH (s)-[r:TRANSFER]->(t:Address)
WHERE r.asset = 'ETH'
RETURN s.address AS source,
       count(r) AS eth_out_tx,
       coalesce(sum(r.amount), 0.0) AS eth_out_amount,
       min(r.time) AS first_tx_time,
       max(r.time) AS last_tx_time,
       count(DISTINCT t) AS unique_recipients")
  > [
  >   {
  >     "source": "0x1db92e2eebc8e0c075a02bea49a2935bcd2dfcf4",
  ... +7 lines (ctrl+r to expand)

- neo4j-gds - read_neo4j_cypher (MCP)(query: "MATCH (s:Address {address: '0x1db92e2eebc8e0c075a02bea49a2935bcd2dfcf4'})-[r:TRANSFER]->(t:Address)
WHERE r.asset = 'ETH'
RETURN t.address AS recipient, r.amount AS amount, r.time AS time, r.tx AS tx
ORDER BY r.amount DESC
LIMIT 10")
  > [
  >   {
  >     "recipient": "0x47666fab8bd0ac7003bce3f5c3585383f09486e2",
  ... +5 lines (ctrl+r to expand)

- neo4j-gds - project_graph_cypher (MCP)(graph_name: "eth_launder_h4", cypher_query: "MATCH (s:Address {address: '0x1db92e2eebc8e0c075a02bea49a2935bcd2dfcf4'})
MATCH p = (s)-[rs:TRANSFER*1..4]->(x:Address)
WHERE ALL(r IN rs WHERE r.asset = 'ETH')
WITH s, collect(DISTINCT x) AS xs
WITH xs + [s] AS nodes
UNWIND nodes AS n
OPTIONAL MATCH (n)-[r:TRANSFER]->(m:Address)
WHERE r.asset = 'ETH' AND m IN nodes
RETURN gds.graph.project(
  $graph_name,
  n, m,
  {
    sourceNodeLabels: labels(n),
    targetNodeLabels: labels(m),
    relationshipType: type(r),
    relationshipProperties: {amount: toFloat(r.amount)}
  }
)")
  > {
  >   "graphName": "eth_launder_h4",
  >   "nodeCount": 7466,
  ... +4 lines (ctrl+r to expand)

- neo4j-gds - pagerank (MCP)(graph_name: "eth_launder_h4", mode: "stream", node_identifier_property: "address", damping_factor: 0.85, relationship_weight_property: "amount", source_nodes: ["0x1db92e2eebc8e0c075a02bea49a2935bcd2dfcf4"], tolerance: 1e-06, max_iterations: 50)
  > Warning: only the first 500 of 7,466 rows are shown. They are in internal node ID order: not sorted by score, not a random sample. The 6,966 rows not shown may contain the most important results. Do not rank, estimate, generalise or pick examples from these rows. For the full result, run the algorithm in mutate mode and query the property with Cypher (gds.graph.nodeProperties.stream with WHERE, ORDER BY and LIMIT).
  >
  >      nodeId     score                                    nodeName
  ... +502 lines (ctrl+r to expand)

- neo4j-gds - read_neo4j_cypher (MCP)(query: "MATCH p=(s:Address {address: '0x1db92e2eebc8e0c075a02bea49a2935bcd2dfcf4'})-[:TRANSFER*1..4]->(:Address)
WHERE ALL(r IN relationships(p) WHERE r.asset='ETH')
UNWIND relationships(p) AS r
WITH DISTINCT r
RETURN percentileCont(r.amount, 0.5) AS p50,
       percentileCont(r.amount, 0.75) AS p75,
       percentileCont(r.amount, 0.9) AS p90,
       percentileCont(r.amount, 0.99) AS p99,
       min(r.amount) AS min_amt,
       max(r.amount) AS max_amt,
       count(DISTINCT r) AS edge_count")
  > [
  >   {
  >     "p50": 53.2668026515573,
  ... +8 lines (ctrl+r to expand)

- neo4j-gds - project_graph_cypher (MCP)(graph_name: "eth_launder_h4_p90", cypher_query: "MATCH (s:Address {address: '0x1db92e2eebc8e0c075a02bea49a2935bcd2dfcf4'})
MATCH p = (s)-[rs:TRANSFER*1..4]->(x:Address)
WHERE ALL(r IN rs WHERE r.asset = 'ETH')
WITH s, collect(DISTINCT x) AS xs
WITH xs + [s] AS nodes
UNWIND nodes AS n
MATCH (n)-[r:TRANSFER]->(m:Address)
WHERE r.asset = 'ETH' AND m IN nodes AND r.amount >= 210.3343885899846
RETURN gds.graph.project(
  $graph_name,
  n, m,
  {
    sourceNodeLabels: labels(n),
    targetNodeLabels: labels(m),
    relationshipType: type(r),
    relationshipProperties: {amount: toFloat(r.amount)}
  }
)")
  > {
  >   "graphName": "eth_launder_h4_p90",
  >   "nodeCount": 1095,
  ... +4 lines (ctrl+r to expand)

- neo4j-gds - project_graph_cypher (MCP)(graph_name: "eth_launder_h4_p99", cypher_query: "MATCH (s:Address {address: '0x1db92e2eebc8e0c075a02bea49a2935bcd2dfcf4'})
MATCH p = (s)-[rs:TRANSFER*1..4]->(x:Address)
WHERE ALL(r IN rs WHERE r.asset = 'ETH')
WITH s, collect(DISTINCT x) AS xs
WITH xs + [s] AS nodes
UNWIND nodes AS n
MATCH (n)-[r:TRANSFER]->(m:Address)
WHERE r.asset = 'ETH' AND m IN nodes AND r.amount >= 311.85351455120934
RETURN gds.graph.project(
  $graph_name,
  n, m,
  {
    sourceNodeLabels: labels(n),
    targetNodeLabels: labels(m),
    relationshipType: type(r),
    relationshipProperties: {amount: toFloat(r.amount)}
  }
)")
  > {
  >   "graphName": "eth_launder_h4_p99",
  >   "nodeCount": 145,
  ... +4 lines (ctrl+r to expand)

- neo4j-gds - pagerank (MCP)(graph_name: "eth_launder_h4_p99", mode: "mutate", mutate_property: "ppr", node_identifier_property: "address", damping_factor: 0.85, relationship_weight_property: "amount", source_nodes: ["0x1db92e2eebc8e0c075a02bea49a2935bcd2dfcf4"], tolerance: 1e-06, max_iterations: 50)
  > ran_iterations=21 did_converge=True centrality_distribution={'p99': 0.1275005331262946, 'min': 0.0, 'max': 0.15000057220458982, 'p90': 0.0027093878015875816, 'mean': 0.0031163101116644925, 'p50': 0.0005709445104002953, 'p999': 0.15000057127326727, 'p95': 0.0027093878015875816, 'p75': 0.0027093878015875816} pre_processing_millis=0 compute_millis=7 post_processing_millis=3 mutate_millis=0 node_properties_written=145 configuration={'jobId': '6ce26018-19d3-4ae7-bb14-5293a0d511f5', 'maxIterations': 50, 'sourceNodes': [202], 'scaler': 'NONE', 'logProgress': True, 'nodeLabels': ['*'], 'sudo': False, 'dampingFactor': 0.85, 'relationshipTypes': ['*'], 'mutateProperty': 'ppr', 'tolerance': 1e-06, 'concurrency': 4}

- neo4j-gds - stream_node_properties (MCP)(graph_name: "eth_launder_h4_p99", node_properties: "ppr", node_labels: "Address", db_node_properties: ["address", "active_before", "tx_count_before"])
  >      nodeId       ppr                                     address  active_before  tx_count_before
  > 0       202  0.150000  0x1db92e2eebc8e0c075a02bea49a2935bcd2dfcf4           True               42
  > 1       473  0.127500  0x47666fab8bd0ac7003bce3f5c3585383f09486e2          False                0
  ... +143 lines (ctrl+r to expand)

- neo4j-gds - HITS (MCP)(graph_name: "eth_launder_h4_p99", mode: "stream", node_identifier_property: "address", hits_iterations: 20)
  >      nodeId                                                           values                                    nodeName
  > 0       473    {'hub': 5.684056874268741e-37, 'auth': 2.539587215782784e-67}  0x47666fab8bd0ac7003bce3f5c3585383f09486e2
  > 1     26068                      {'hub': 0.0, 'auth': 7.242532913748323e-37}  0x09278b36863be4ccd3d0c22d643e8062d7a11377
  ... +143 lines (ctrl+r to expand)

- neo4j-gds - betweenness_centrality (MCP)(graph_name: "eth_launder_h4_p99", mode: "stream", node_identifier_property: "address")
  >      nodeId       score                                    nodeName
  > 0       473  104.000000  0x47666fab8bd0ac7003bce3f5c3585383f09486e2
  > 1     26068    0.000000  0x09278b36863be4ccd3d0c22d643e8062d7a11377
  ... +143 lines (ctrl+r to expand)

- neo4j-gds - pagerank (MCP)(graph_name: "eth_launder_h4_p90", mode: "stream", node_identifier_property: "address", damping_factor: 0.85, relationship_weight_property: "amount", source_nodes: ["0x1db92e2eebc8e0c075a02bea49a2935bcd2dfcf4"], tolerance: 1e-06, max_iterations: 50, nodes: [21 candidate addresses])
  >      nodeId     score                                    nodeName
  > 0       473  0.127500  0x47666fab8bd0ac7003bce3f5c3585383f09486e2
  > 3     40013  0.002306  0x140c9ab92347734641b1a7c124ffdee58c20c3e3
  ... +19 lines (ctrl+r to expand)

- neo4j-gds - betweenness_centrality (MCP)(graph_name: "eth_launder_h4_p90", mode: "stream", node_identifier_property: "address", nodes: [21 candidate addresses])
  >      nodeId         score                                    nodeName
  > 0       473   1015.000000  0x47666fab8bd0ac7003bce3f5c3585383f09486e2
  > 3     40013     94.071203  0x140c9ab92347734641b1a7c124ffdee58c20c3e3
  ... +19 lines (ctrl+r to expand)

- neo4j-gds - read_neo4j_cypher (MCP)(query: "UNWIND [21 candidate addresses] AS a
MATCH (n:Address {address: a})
RETURN n.address AS address, n.active_before AS active_before, n.tx_count_before AS tx_count_before")
  > [
  >   {
  >     "address": "0x47666fab8bd0ac7003bce3f5c3585383f09486e2",
  ... +104 lines (ctrl+r to expand)

- neo4j-gds - read_neo4j_cypher (MCP)(query: "MATCH (s:Address {address: '0x47666fab8bd0ac7003bce3f5c3585383f09486e2'})-[r:TRANSFER]->(t:Address)
WHERE r.asset='ETH'
RETURN t.address AS to, sum(r.amount) AS eth_received, count(r) AS txs
ORDER BY eth_received DESC
LIMIT 50")
  > [
  >   {
  >     "to": "0xb4a862a81abb2f952fca4c6f5510962e18c7f1a2",
  ... +249 lines (ctrl+r to expand)

- neo4j-gds - read_neo4j_cypher (MCP)(query: "MATCH (a:Address)-[r:TRANSFER]->(b:Address {address: '0xd37bbe5744d730a1d98d8dc97c42f0ca46ad7146'})
WHERE r.asset='ETH'
RETURN a.address AS from_addr, sum(r.amount) AS amount, count(r) AS txs
ORDER BY amount DESC LIMIT 20")
  > [
  >   {
  >     "from_addr": "0x372a95e96f22ed808961ef8ae32fe0873758c71b",
  ... +99 lines (ctrl+r to expand)

- Assumptions and checks
  - Verified the graph schema and fields with GDS schema introspection. Nodes: Address(address, active_before, tx_count_before), relationships: TRANSFER(asset, amount, time, tx, ...).
  - Confirmed the victim 0x1db92e2eebc8e0c075a02bea49a2935bcd2dfcf4 sent 401,346.7689 ETH in a single ETH transfer to 0x47666fab8bd0ac7003bce3f5c3585383f09486e2.
  - Projected a 4-hop ETH-only subgraph from the victim. Size without filtering: 7,466 addresses, 46,925 edges. Measured transfer-amount percentiles across those edges: p50~53.27, p75~107.87, p90~210.33, p99~311.85 ETH.
  - Method 1 (amount-weighted Personalized PageRank, PPR): projected a value-filtered subgraph at p99 (>=311.85 ETH), ran PPR personalized to the victim (weighted by amount). This highlights high-probability downstream recipients and aggregators while suppressing dust. Most high scorers were "fresh" addresses (active_before=false).
  - Method 2 (bridge/flow structure): on a broader p90 graph (>=210.33 ETH), ran betweenness centrality to identify bridge/mixing nodes that sit on many shortest routes of the flow, then cross-checked with direct outflows of 0x4766... to enumerate the immediate 10,000 ETH splits. This method independently singled out the same pivot and mixer nodes.
  - Sensitivity: Re-ran PPR on the broader p90 graph but restricted the result readout to the Method-1 candidates. Absolute PPR scores shifted (e.g., 0x528292... fell from ~0.0053 to ~0.00034 as mass spread to more edges), but the top addresses by both PPR and betweenness remained in the leading group: 0x4766... (pivot), direct 10k recipients, and mixing funnels such as 0x528292..., 0x1fa386..., 0x344ac3..., 0xa1198f..., 0x023b29....
  - Quality controls: Excluded obvious infrastructure with extreme tx_count_before (e.g., WETH 0xc02a..., large router 0xba3c...), unless they are uniquely central in the victim's subgraph. One such hub, 0xd37bbe..., is extremely central in our subgraph; we keep it but rank it lower than the fresh laundering cluster.
  - Least-sure items: Heavily reused global routers (e.g., 0xd37bbe...) can appear due to background network traffic; we included only those that remained central and connected within the victim's filtered, 4-hop, high-value flows, and we placed them after the direct 10k split recipients.

  ANSWER
  0x47666fab8bd0ac7003bce3f5c3585383f09486e2
  0xb4a862a81abb2f952fca4c6f5510962e18c7f1a2
  0x1eb27f136bfe7947f80d6cee3cf0bfdf92b45e57
  0xb72334cb9d0b614d30c4c60e2bd12ff5ed03c305
  0xbca02b395747d62626a65016f2e64a20bd254a39
  ... (59 more addresses, 64 in total)
\end{lstlisting}
\captionof{figure}{Full transcript of the Bybit run in Figure~\ref{lst:bybit} (GPT-5, high effort).}
\label{lst:bybit_full}

Figure~\ref{lst:bybit_failed} shows a run in which the agent performed poorly (F1 = 0.02, precision 0.02, recall 0.02; 1 of 60 returned addresses is correct). After its GDS calls failed, the agent computed two Cypher-based rankings, trusted the one based on net inflow and excluded addresses with high onward outflow as ``intermediaries''. The other ranking, which it discarded, would have contained 40 of the 51 true addresses in its top 60 (precision 0.67, recall 0.78).

\begin{lstlisting}
> You are a senior data scientist. You have a database of raw Ethereum transfers in a time window. The data is raw and may contain noise, misleading activity and quality issues. Funds were stolen from 0x1db92e2eebc8e0c075a02bea49a2935bcd2dfcf4. Using GDS algorithms, identify the addresses used to launder them. Do not stop at your first result. Produce your list in two different ways and compare them, test how much it changes when you change your filters or thresholds, and inspect the addresses you are least sure about. If your variants disagree, say which you trust and why, and build your answer from that one. Wrong answers and missed real addresses both count against you. Return a ranked list of up to 100 addresses, most confident first. Include only addresses you can justify; fewer is fine. Answer format: first a short summary of your assumptions, the checks you ran, and what you changed because of them. Then the heading ANSWER, followed by the addresses, one per line. Write nothing after the list.

- neo4j-gds - get_neo4j_schema (MCP)()
  > {
  >   "Address": {
  ... +118 lines (ctrl+r to expand)

- neo4j-gds - project_graph_cypher (MCP)(graph_name: "eth_theft_subgraph", cypher_query: "MATCH (s:Address {address: '0x1db92e2eebc8e0c075a02bea49a2935bcd2dfcf4'})
CALL {
  WITH s
  MATCH path = (s)-[:TRANSFER*1..4]->(x:Address)
  WHERE ALL(r IN relationships(path) WHERE r.asset = 'ETH')
  RETURN collect(DISTINCT x) + s AS nodes
}
WITH nodes
UNWIND nodes AS n
UNWIND nodes AS m
MATCH (n)-[r:TRANSFER]->(m)
WHERE r.asset = 'ETH' AND n IN nodes AND m IN nodes
RETURN gds.graph.project(
 $graph_name, n, m,
 {
   sourceNodeLabels: labels(n),
   targetNodeLabels: labels(m),
   sourceNodeProperties: {tx_count_before: toFloat(n.tx_count_before)},
   targetNodeProperties: {tx_count_before: toFloat(m.tx_count_before)},
   relationshipType: type(r),
   relationshipProperties: {
     amount: toFloat(r.amount),
     ts: timestamp(r.time)
   }
 }
)")
  > Error: {neo4j_code: Neo.ClientError.Statement.SyntaxError} {message: Aggregation column contains implicit grouping expressions. For example, in 'RETURN n.a, n.a + n.b + count(*)' the aggregation expression 'n.a + n.b + count(*)' includes the implicit grouping key 'n.b'. It may be possible to rewrite the query by extracting these grouping/aggregation expressions into a preceding WITH clause. Illegal expression(s): s (line 6, column 32 (offset: 230))
  > "  RETURN collect(DISTINCT x) + s AS nodes"
  >                                 ^} {gql_status: 42001} {gql_status_description: error: syntax error or access rule violation - invalid syntax}

- neo4j-gds - project_graph_cypher (MCP)(graph_name: "eth_theft_subgraph", cypher_query: "MATCH (s:Address {address: '0x1db92e2eebc8e0c075a02bea49a2935bcd2dfcf4'})
CALL {
  WITH s
  MATCH path = (s)-[:TRANSFER*1..4]->(x:Address)
  WHERE ALL(r IN relationships(path) WHERE r.asset = 'ETH')
  RETURN collect(DISTINCT x) AS xs, s
}
WITH xs + [s] AS nodes
UNWIND nodes AS n
UNWIND nodes AS m
MATCH (n)-[r:TRANSFER]->(m)
WHERE r.asset = 'ETH' AND n IN nodes AND m IN nodes
RETURN gds.graph.project(
 $graph_name, n, m,
 {
   sourceNodeLabels: labels(n),
   targetNodeLabels: labels(m),
   sourceNodeProperties: {tx_count_before: toFloat(n.tx_count_before)},
   targetNodeProperties: {tx_count_before: toFloat(m.tx_count_before)},
   relationshipType: type(r),
   relationshipProperties: {
     amount: toFloat(r.amount),
     ts: timestamp(r.time)
   }
 }
)")
  > Error: {neo4j_code: Neo.ClientError.Statement.SyntaxError} {message: Variable `s` already declared in outer scope (line 6, column 37 (offset: 235))
  > "  RETURN collect(DISTINCT x) AS xs, s"
  >                                      ^} {gql_status: 42001} {gql_status_description: error: syntax error or access rule violation - invalid syntax}

- neo4j-gds - project_graph_cypher (MCP)(graph_name: "eth_theft_subgraph", cypher_query: "MATCH (s:Address {address: '0x1db92e2eebc8e0c075a02bea49a2935bcd2dfcf4'})
CALL {
  WITH s
  MATCH path = (s)-[:TRANSFER*1..4]->(x:Address)
  WHERE ALL(r IN relationships(path) WHERE r.asset = 'ETH')
  RETURN collect(DISTINCT x) AS xs
}
WITH xs, s
WITH xs + [s] AS nodes
UNWIND nodes AS n
UNWIND nodes AS m
MATCH (n)-[r:TRANSFER]->(m)
WHERE r.asset = 'ETH' AND n IN nodes AND m IN nodes
RETURN gds.graph.project(
 $graph_name, n, m,
 {
   sourceNodeLabels: labels(n),
   targetNodeLabels: labels(m),
   sourceNodeProperties: {tx_count_before: toFloat(n.tx_count_before)},
   targetNodeProperties: {tx_count_before: toFloat(m.tx_count_before)},
   relationshipType: type(r),
   relationshipProperties: {
     amount: toFloat(r.amount),
     ts: timestamp(r.time)
   }
 }
)")
  > {
  >   "graphName": "eth_theft_subgraph",
  ... +5 lines (ctrl+r to expand)

- neo4j-gds - article_rank (MCP)(graph_name: "eth_theft_subgraph", mode: "stream", node_identifier_property: "address", source_nodes: ["0x1db92e2eebc8e0c075a02bea49a2935bcd2dfcf4"], damping_factor: 0.85, max_iterations: 50, tolerance: 1e-06)
  > Error: {neo4j_code: Neo.ClientError.Procedure.ProcedureCallFailed} {message: Failed to invoke procedure `gds.articleRank.stream`: Caused by: org.neo4j.gds.memory.tracking.AvailableMemoryReservationExceededException: Memory required to run ArticleRank (180944b) exceeds current available memory (-1410789793643b).} {gql_status: 52N37} {gql_status_description: error: procedure exception - procedure execution error. Execution of the procedure gds.articleRank.stream() failed.}

- neo4j-gds - drop_graph (MCP)(graph_name: "eth_theft_subgraph")
  > {
  >   "graphName": "eth_theft_subgraph",
  ... +3 lines (ctrl+r to expand)

- neo4j-gds - project_graph_cypher (MCP)(graph_name: "eth_theft_subgraph", cypher_query: "MATCH (s:Address {address: '0x1db92e2eebc8e0c075a02bea49a2935bcd2dfcf4'})
CALL {
  WITH s
  MATCH path = (s)-[rp:TRANSFER*1..3]->(x:Address)
  WHERE ALL(r IN rp WHERE r.asset = 'ETH' AND r.amount >= 0.001)
  RETURN collect(DISTINCT x) AS xs
}
WITH xs, s
WITH xs + [s] AS nodes
UNWIND nodes AS n
UNWIND nodes AS m
MATCH (n)-[r:TRANSFER]->(m)
WHERE r.asset = 'ETH' AND r.amount >= 0.001 AND n IN nodes AND m IN nodes
RETURN gds.graph.project(
 $graph_name, n, m,
 {
   sourceNodeLabels: labels(n),
   targetNodeLabels: labels(m),
   sourceNodeProperties: {tx_count_before: toFloat(n.tx_count_before)},
   targetNodeProperties: {tx_count_before: toFloat(m.tx_count_before)},
   relationshipType: type(r),
   relationshipProperties: {
     amount: toFloat(r.amount),
     ts: timestamp(r.time)
   }
 }
)")
  > {
  >   "graphName": "eth_theft_subgraph",
  ... +5 lines (ctrl+r to expand)

- neo4j-gds - article_rank (MCP)(graph_name: "eth_theft_subgraph", mode: "stream", node_identifier_property: "address", source_nodes: ["0x1db92e2eebc8e0c075a02bea49a2935bcd2dfcf4"], damping_factor: 0.85, max_iterations: 50, tolerance: 1e-06)
  > Error: {neo4j_code: Neo.ClientError.Procedure.ProcedureCallFailed} {message: Failed to invoke procedure `gds.articleRank.stream`: Caused by: org.neo4j.gds.memory.tracking.AvailableMemoryReservationExceededException: Memory required to run ArticleRank (70816b) exceeds current available memory (-1410789742599b).} {gql_status: 52N37} {gql_status_description: error: procedure exception - procedure execution error. Execution of the procedure gds.articleRank.stream() failed.}

- neo4j-gds - degree_centrality (MCP)(graph_name: "eth_theft_subgraph", mode: "stream", node_identifier_property: "address", orientation: "NATURAL")
  > Error: {neo4j_code: Neo.ClientError.Procedure.ProcedureCallFailed} {message: Failed to invoke procedure `gds.degree.stream`: Caused by: org.neo4j.gds.memory.tracking.AvailableMemoryReservationExceededException: Memory required to run DegreeCentrality (72b) exceeds current available memory (-1410789742599b).} {gql_status: 52N37} {gql_status_description: error: procedure exception - procedure execution error. Execution of the procedure gds.degree.stream() failed.}

- neo4j-gds - read_neo4j_cypher (MCP)(query: "MATCH (s:Address {address: '0x1db92e2eebc8e0c075a02bea49a2935bcd2dfcf4'})
MATCH p = (s)-[rs:TRANSFER*1..3]->(t:Address)
WHERE ALL(r IN rs WHERE r.asset = 'ETH' AND r.amount >= 0.001)
WITH t, length(p) AS L, last(rs) AS lr
WITH t, L, lr, coalesce(toFloat(t.tx_count_before),0) AS txc
WITH t.address AS address, sum(lr.amount / L / (1 + log10(txc + 1))) AS score
RETURN address, score
ORDER BY score DESC
LIMIT 300")
  > [
  >   {
  ... +1200 lines (ctrl+r to expand)

- neo4j-gds - read_neo4j_cypher (MCP)(query: "MATCH (s:Address {address: '0x1db92e2eebc8e0c075a02bea49a2935bcd2dfcf4'})
CALL {
  WITH s
  MATCH p = (s)-[rs:TRANSFER*1..3]->(t:Address)
  WHERE ALL(r IN rs WHERE r.asset = 'ETH' AND r.amount >= 0.001)
  WITH t, min(length(p)) AS d
  RETURN collect({node: t, d: d}) AS targets, collect(DISTINCT t) + [s] AS nodes
}
WITH targets, nodes
UNWIND targets AS tinfo
WITH tinfo.node AS t, tinfo.d AS d, nodes
OPTIONAL MATCH (src:Address)-[rin:TRANSFER]->(t)
WHERE rin.asset = 'ETH' AND rin.amount >= 0.001 AND src IN nodes
WITH t, d, sum(rin.amount) AS in_amt, nodes
OPTIONAL MATCH (t)-[rout:TRANSFER]->(dst:Address)
WHERE rout.asset = 'ETH' AND rout.amount >= 0.001 AND dst IN nodes
WITH t, d, in_amt, coalesce(sum(rout.amount),0) AS out_amt
WITH t.address AS address, d, in_amt, out_amt, (in_amt - out_amt) AS net
WHERE in_amt > 0
RETURN address, net, d AS hops, (net / (1 + hops)) AS score
ORDER BY score DESC
LIMIT 300")
  > Error: {neo4j_code: Neo.ClientError.Statement.SyntaxError} {message: Variable `s` not defined (line 7, column 70 (offset: 302))
  > "  RETURN collect({node: t, d: d}) AS targets, collect(DISTINCT t) + [s] AS nodes"
  >                                                                       ^} {gql_status: 42001} {gql_status_description: error: syntax error or access rule violation - invalid syntax}

- neo4j-gds - read_neo4j_cypher (MCP)(query: "MATCH (s:Address {address: '0x1db92e2eebc8e0c075a02bea49a2935bcd2dfcf4'})
CALL {
  WITH s
  MATCH p = (s)-[rs:TRANSFER*1..3]->(t:Address)
  WHERE ALL(r IN rs WHERE r.asset = 'ETH' AND r.amount >= 0.001)
  WITH t, min(length(p)) AS d
  RETURN collect({node: t, d: d}) AS targets, collect(DISTINCT t) AS ns
}
WITH targets, ns, s
WITH targets, ns + [s] AS nodes
UNWIND targets AS tinfo
WITH tinfo.node AS t, tinfo.d AS d, nodes
OPTIONAL MATCH (src:Address)-[rin:TRANSFER]->(t)
WHERE rin.asset = 'ETH' AND rin.amount >= 0.001 AND src IN nodes
WITH t, d, sum(rin.amount) AS in_amt, nodes
OPTIONAL MATCH (t)-[rout:TRANSFER]->(dst:Address)
WHERE rout.asset = 'ETH' AND rout.amount >= 0.001 AND dst IN nodes
WITH t, d, in_amt, coalesce(sum(rout.amount),0) AS out_amt
WITH t.address AS address, d AS hops, in_amt, out_amt, (in_amt - out_amt) AS net
WHERE in_amt > 0
RETURN address, net, hops, (net / (1 + hops)) AS score
ORDER BY score DESC
LIMIT 300")
  > [
  >   {
  ... +1800 lines (ctrl+r to expand)

- neo4j-gds - read_neo4j_cypher (MCP)(query: "MATCH (s:Address {address: '0x1db92e2eebc8e0c075a02bea49a2935bcd2dfcf4'})
MATCH p = (s)-[rs:TRANSFER*1..3]->(t:Address)
WHERE ALL(r IN rs WHERE r.asset = 'ETH' AND r.amount >= 0.01)
WITH t, length(p) AS L, last(rs) AS lr
WITH t, L, lr, coalesce(toFloat(t.tx_count_before),0) AS txc
WITH t.address AS address, sum(lr.amount / L / (1 + log10(txc + 1))) AS score
RETURN address, score
ORDER BY score DESC
LIMIT 200")
  > [
  >   {
  ... +800 lines (ctrl+r to expand)

- neo4j-gds - read_neo4j_cypher (MCP)(query: "MATCH (s:Address {address: '0x1db92e2eebc8e0c075a02bea49a2935bcd2dfcf4'})
MATCH p = (s)-[rs:TRANSFER*1..2]->(t:Address)
WHERE ALL(r IN rs WHERE r.asset = 'ETH' AND r.amount >= 0.001)
WITH t, length(p) AS L, last(rs) AS lr
WITH t, L, lr, coalesce(toFloat(t.tx_count_before),0) AS txc
WITH t.address AS address, sum(lr.amount / L / (1 + log10(txc + 1))) AS score
RETURN address, score
ORDER BY score DESC
LIMIT 200")
  > [
  >   {
  ... +248 lines (ctrl+r to expand)

- neo4j-gds - read_neo4j_cypher (MCP)(query: "MATCH (s:Address {address: '0x1db92e2eebc8e0c075a02bea49a2935bcd2dfcf4'})
CALL {
  WITH s
  MATCH p = (s)-[rs:TRANSFER*1..3]->(t:Address)
  WHERE ALL(r IN rs WHERE r.asset = 'ETH' AND r.amount >= 0.01)
  WITH t, min(length(p)) AS d
  RETURN collect({node: t, d: d}) AS targets, collect(DISTINCT t) AS ns
}
WITH targets, ns, s
WITH targets, ns + [s] AS nodes
UNWIND targets AS tinfo
WITH tinfo.node AS t, tinfo.d AS d, nodes
OPTIONAL MATCH (src:Address)-[rin:TRANSFER]->(t)
WHERE rin.asset = 'ETH' AND rin.amount >= 0.01 AND src IN nodes
WITH t, d, sum(rin.amount) AS in_amt, nodes
OPTIONAL MATCH (t)-[rout:TRANSFER]->(dst:Address)
WHERE rout.asset = 'ETH' AND rout.amount >= 0.01 AND dst IN nodes
WITH t, d, in_amt, coalesce(sum(rout.amount),0) AS out_amt
WITH t.address AS address, d AS hops, in_amt, out_amt, (in_amt - out_amt) AS net
WHERE in_amt > 0
RETURN address, net, hops, (net / (1 + hops)) AS score
ORDER BY score DESC
LIMIT 150")
  > [
  >   {
  ... +900 lines (ctrl+r to expand)

- Assumptions and checks:
  - Schema inspection confirmed Address nodes and TRANSFER relationships with amount (float), asset (string), and time. I restricted to ETH transfers.
  - I projected a 3-hop subgraph from the theft source 0x1db92e2eebc8e0c075a02bea49a2935bcd2dfcf4 with GDS (7,466 reduced to 2,901 nodes; amount >= 0.001 to drop dust). GDS algorithm calls (ArticleRank, Degree, etc.) failed with memory reservation errors, so I computed two Cypher-based taint rankings on that projected neighborhood as an estimate, per troubleshooting guidance.
  - Method A (path-weighted sink score): sum over all paths of last-hop amount, discounted by hop length and pre-theft activity log penalty; Method B (net-inflow sinks): net incoming - outgoing ETH within the 3-hop neighborhood, divided by (1 + hops).
  - Robustness: I varied filters (amount >= 0.001 vs 0.01; depth 2 vs 3). The top sinks were stable, notably 0xdd9007 ... remained the dominant sink. Addresses that were strong in Method A but weak in Method B (e.g., 0x47666f ...) had high onward outflow, so I treat them as intermediaries and exclude them from the final laundering list.
  - I trust Method B more for identifying laundering deposit addresses because it rewards accumulation (net inflow) rather than mere proximity or hub effects. The final ranked list is therefore built from Method B, cross-checked to appear in Method A's high ranks where possible. I excluded addresses that dropped out under stricter thresholds or showed strong onward distribution.

  ANSWER
  0xdd90071d52f20e85c89802e5dc1ec0a7b6475f92
  0x4571bd67d14280e40bf3910bd39fbf60834f900a
  0x1f9b4d915a6d2c65251d72522f297904bf6887ce
  0x403bae4f3f9f87e057f4380e6f549199673c9d13
  0x8ed8553dd6375766d712c1f296f2afcface96036
  ... (55 more addresses, 60 in total)
\end{lstlisting}
\captionof{figure}{A Bybit run with low recall (F1 = 0.02, precision 0.02, recall 0.02).}
\label{lst:bybit_failed}


\end{document}